\documentclass[10pt,twocolumn,letterpaper]{article}

\usepackage{cvpr}
\usepackage{times}
\usepackage{epsfig}
\usepackage{graphicx}
\usepackage{amsmath}
\usepackage{bbold}
\usepackage{amssymb}
\usepackage{pifont}
\usepackage{bbm}
\usepackage{caption}
\usepackage{subcaption}

\usepackage[dvipsnames]{xcolor}
\usepackage{xspace}

\newcommand{\sota}{state of the art }
\newcommand{\icl}{ICL}
\newcommand{\ours}{MiB }
\newcommand{\expandednick}{\textbf{M}odel\textbf{i}ng the \textbf{B}ackground for incremental learning in semantic segmentation}

\newcommand{\hypers}{hyper-parameters }
\newcommand{\hyper}{hyper-parameter }
\newcommand{\set}{\mathcal}
\newcommand{\con}{\mathtt}
\DeclareMathOperator*{\argmax}{arg\,max} % Jan Hlavacek
\newcommand{\real}{{\rm I\!R}}

\newcommand{\myparagraph}[1]{\vspace{4pt}\noindent\textbf{#1}}

\usepackage[pagebackref=true,breaklinks=true,colorlinks,bookmarks=false]{hyperref}

\cvprfinalcopy % *** Uncomment this line for the final submission

\def\cvprPaperID{5832} % *** Enter the CVPR Paper ID here

\ifcvprfinal\fi
\begin{document}

%%%%%%%%% TITLE
\title{Modeling the Background for Incremental Learning in Semantic Segmentation}

\author{
Fabio Cermelli$^{1,2}$, Massimiliano Mancini$^{2,3,4}$, Samuel Rota Bul\`o$^{5}$, Elisa Ricci$^{3,6}$, Barbara Caputo$^{1,2}$\\
$^1$Politecnico di Torino, $^2$Italian Institute of Technology, $^3$Fondazione Bruno Kessler,\\ $^4$Sapienza University of Rome,
$^5$Mapillary Research, $^6$University of Trento \\
{\tt\small \{fabio.cermelli, barbara.caputo\}@polito.it, mancini@diag.uniroma1.it,} \\
{\tt\small samuel@mapillary.com, eliricci@fbk.eu}
}

\maketitle

\begin{abstract}
% WHY
%Deep learning has enabled enormous progress in computer vision. 
Despite their effectiveness in a wide range of tasks, deep architectures suffer from some important limitations. In particular, 
%one main weakness of neural networks is that 
they are 
%especially 
vulnerable to catastrophic forgetting, \ie they perform poorly when they are required to update their model as new classes are available but the original training set is not retained. % in absence of original training data and annotations of original classes in the new training set.
This paper addresses this problem in the context of semantic segmentation. Current 
%incremental learning 
strategies fail on this task because they do not consider a peculiar aspect of semantic segmentation: since each training step provides annotation only for a subset of all possible classes, pixels of the background class (\ie pixels that do not belong to any other classes) exhibit a semantic distribution shift.
% WHAT % importance to model background 
In this work we revisit classical incremental learning methods, proposing a new distillation-based framework which explicitly accounts for this shift.
%{In this work we propose a novel distillation-based framework for incremental learning which explicitly accounts for this shift through appropriately designed loss functions.}  %classification and knowledge distillation losses commonly adopted in previous methods to explicitly solve this problem.
Furthermore, we introduce a novel strategy to initialize classifier's parameters, thus preventing %to 
biased predictions toward the background class. % Results
We demonstrate the effectiveness of our approach with an extensive evaluation on the Pascal-VOC 2012 and ADE20K datasets, 
%showing that our approach 
significantly outperforming \sota incremental learning methods. %that we adapted for the semantic segmentation task. 
% \fabio{Dobbiamo rendere più esplcito che siamo tra i primi ad afforntare il problema di ICL in SS?}
Code can be found at \url{https://github.com/fcdl94/MiB}.
\end{abstract}

\section{Introduction}

\begin{figure}[t]
    \centering
    \includegraphics[width=\linewidth]{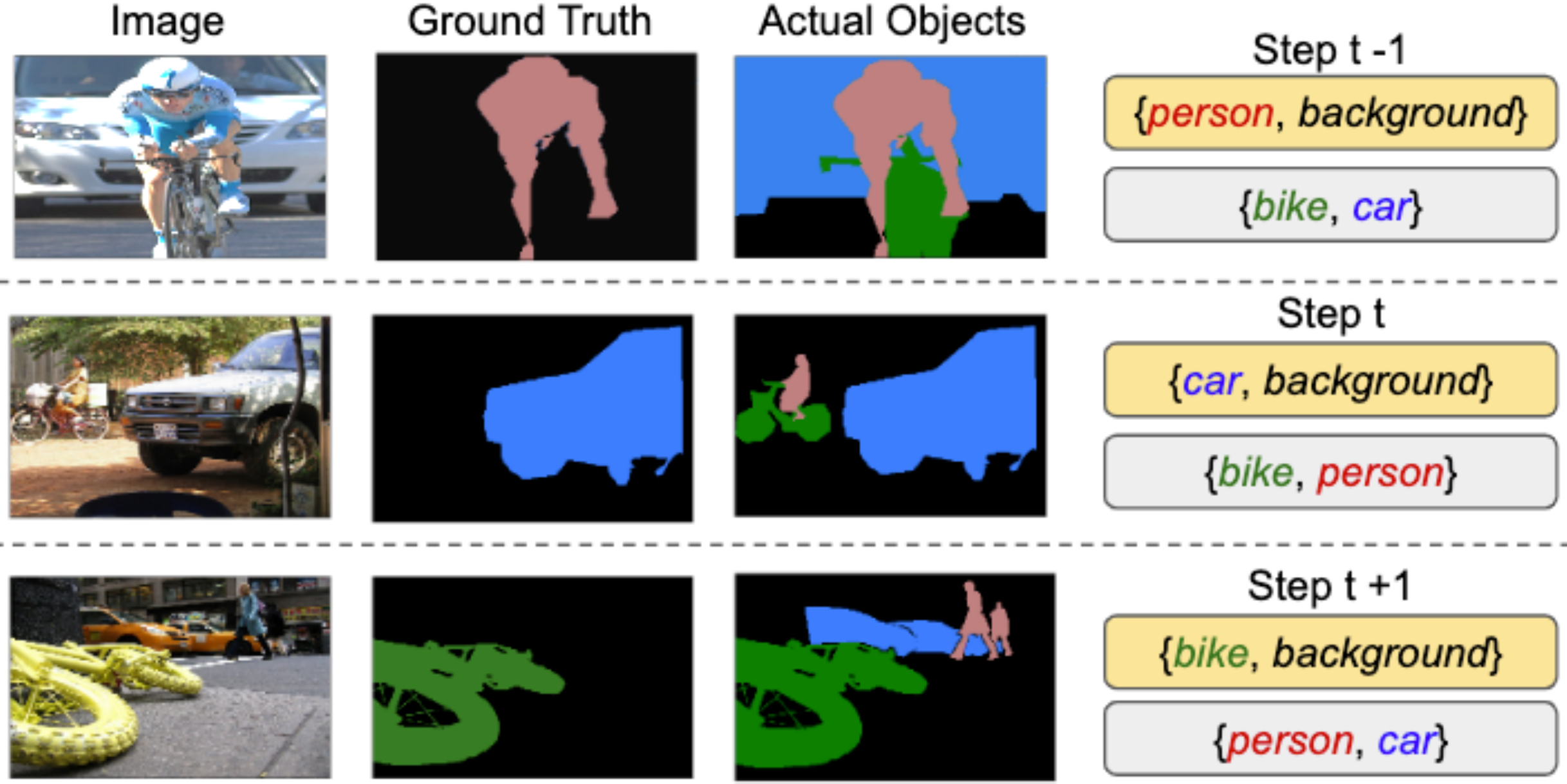}
    \vspace{-10pt}
    \caption{Illustration of the semantic shift of the background class in incremental learning for semantic segmentation. Yellow boxes denote the ground truth provided in the learning step, while grey boxes denote classes not labeled. As different learning steps have different label spaces, at step $t$ old classes (\eg \textit{person}) and unseen ones (\eg \textit{car}) might be labeled as background in the current ground truth. % (\eg \textit{person}). %of current and future learning steps (\eg $t$ and $t+1$). 
    %Similarly, also unseen classes might be labeled as background in previous learning steps (\eg \textit{bike}). 
    %While the figure shows 
    Here we  show the specific case of single class learning steps, but we address the general case where an arbitrary number of classes is added.} %the ground truth background contains \textit{car} and \textit{bike} classes. In step 2 it contains the already learned \textit{person} and \textit{bike} classes, while in step 3 it includes \textit{car} and \textit{person}, which are already learned classes.  }
    %\eli{dire cosa è grigio e arancio}
    \label{fig:teaser}
    \vspace{-15pt}
\end{figure}

% Progresses on SS

% Why ICL is important

% Peculiarities

% Contribution

Semantic segmentation
%, \ie the task of inferring for every pixel in an image the semantic category of the object to which it belongs, 
is a fundamental problem in computer vision.  
In the last years, thanks to the emergence of deep neural networks and to the availability of large-scale human-annotated datasets \cite{pascal-voc-2012,zhou2017scene}, the state of the art %in semantic segmentation 
has improved significantly \cite{long2015fully, chen2018encoder, zhao2017pyramid, lin2017refinenet, zhang2018exfuse}.
Current approaches %for semantic segmentation 
are derived by extending deep architectures from image-level to pixel-level classification, taking advantage of Fully Convolutional Networks (FCNs) \cite{long2015fully}. Over the years, semantic segmentation models based on FCNs have been improved in several ways, \eg by exploiting multiscale representations \cite{lin2017refinenet,zhang2018exfuse}, modeling spatial dependencies and contextual cues \cite{chen2017rethinking, chen2017deeplab, chen2018encoder} or considering attention models \cite{chen2016attention}.

%Despite the advances
Still, existing semantic segmentation methods %still lack one crucial ability, \ie they 
are not designed 
%in order 
to incrementally update their inner classification model when new categories are discovered. 
%In general, 
While deep nets are undoubtedly powerful, it is well known that their capabilities in an incremental learning setting are limited \cite{kemker2018measuring}. In fact, deep architectures struggle in updating their parameters for learning new categories whilst preserving good performance on the old ones
%. In other words, they are prone to 
(\textit{catastrophic forgetting} \cite{mccloskey1989catastrophic}).

While the problem of incremental learning has been traditionally addressed in 
%the context of image classification for 
object recognition \cite{li2017learning,kirkpatrick2017overcoming,chaudhry2018riemannian,rebuffi2017icarl,hou2019learning} and detection \cite{shmelkov2017incremental}, much less attention has been devoted to semantic segmentation.
%In this work, 
Here we fill this gap, 
%and we 
proposing an incremental class learning (\icl) approach for semantic segmentation. Inspired by previous methods on image classification \cite{li2017learning,rebuffi2017icarl,castro2018end}, we 
%propose to 
cope with catastrophic forgetting by resorting to knowledge distillation \cite{hinton2015distilling}. However, 
%in this paper 
we argue (and experimentally demonstrate) that a naive application of previous knowledge distillation strategies would not suffice in this setting. In fact, one peculiar aspect of semantic segmentation is the presence of a special class, the background class, indicating pixels not assigned to any of the given object categories. While the presence of this class marginally influences the design of traditional, offline semantic segmentation methods, this is not true in an incremental learning setting.
As illustrated in Fig. \ref{fig:teaser}, it is reasonable to assume that the semantics associated to the background class changes over time. In other words, pixels associated to the background during a learning step may be assigned to a specific object class in subsequent steps or vice-versa, with the effect of exacerbating the catastrophic forgetting. %\eli{spiegare bene nella caption} %In fact, different learning steps contain different label spaces, thus different unlabeled pixels in the ground-truth (\eg Figure \ref{fig:teaser}). 
%This aspect makes even more severe the catastrophic forgetting problem, since pixels of old classes could be labeled as backgrounds in future learning steps and vice-versa. Here we turn this issue into a strength,
%To overcome this issue and properly model the semantic distribution shift within the background class, in this paper we propose to revisit the cross-entropy and the distillation losses adopted in previous \icl\ methods \cite{li2017learning} and introduce the first \icl\ approach for semantic segmentation. 
%\fabio{To overcome this issue, we propose a novel distillation-based framework which refines the cross-entropy and the distillation losses adopted in previous \icl\ methods \cite{li2017learning} to properly model the semantic distribution shift within the background class, introducing the first \icl\ approach for semantic segmentation.} 
To overcome this issue, we revisit the classical distillation-based framework for incremental learning \cite{li2017learning} by introducing two novel loss terms 
%in order 
to properly account for the semantic distribution shift within the background class, thus introducing the first \icl\ approach tailored to semantic segmentation. %\eli{forse la frase precedente da fare + forte} 
We extensively evaluate our method on two datasets, Pascal-VOC \cite{pascal-voc-2012} and ADE20K \cite{zhou2017scene}, showing that our approach, coupled with a novel classifier initialization strategy, largely outperform traditional \icl\ methods. % and previous methods tackling \icl\ in semantic segmentation \cite{michieli2019incremental} in different settings. 

%Differently from standard \icl\ settings considered for image classification,  in semantic segmentation we have that two different incremental learning steps share the void/background class. However,it is reasonable to assume that the distribution of the back-ground class is changing across different incremental steps(see  Fig.   to  illustrate  the  intuition).   In  fact,  backgroundpixels in Yt 1could have been assigned to a class in Ctanndvice-versa.  In the following, we show how we can accountfor the semantic shift in the background class by modifying the supervised cross-entropy (Eq.(2)) and the distilla-tion loss (Eq.(3)) and, thus, how we effectively tackle theincremental class learning problem in semantic segmenta-tion.  We also describe an effective strategy for initializing,at each learning stept, the parameters of the classifier forthe newly introduced classes
%We evaluate the proposed method on two datasets...

\noindent To summarize, the contributions of this paper are as follows:
\vspace{-0.4cm}
\begin{itemize}
    \item We study the  task of incremental class learning for semantic segmentation, analyzing in particular the problem of distribution shift arising due to the presence of the background class. %\massi{diciamo da qualche parte di quelli di Padova/differenze?}
    \vspace{-0.2cm}
    \item We propose a new objective function 
    %for training our deep network for \icl\ 
    and introduce a specific classifier initialization strategy to explicitly cope with the evolving semantics of the background class. We show that our approach greatly alleviates the catastrophic forgetting, leading to the state of the art. %superior performance with respect to baseline methods. %largely outperforms all the baselines in all benchmarks, showing the effectiveness of our approach.
\vspace{-0.2cm}
    \item We benchmark our approach over several previous \icl\ methods on two popular semantic segmentation datasets, considering different experimental settings. We hope that our results will serve as a reference for future works.% and this paper would encourage research on this novel and relevant topic. 
     
\end{itemize}

%In the following we first review the state of the art in incremental learning and semantic segmentation (Sec. \ref{sec:related}), we then formalize the problem and present our approach (Sec. \ref{sec:method}) and finally, we show the benchmarks and experimental analysis (Sec. \ref{sec:exp}).

\section{Related Works}
\label{sec:related}
\myparagraph{Semantic Segmentation.}
%In recent years, 
Deep learning has enabled great advancements in semantic segmentation \cite{long2015fully, chen2018encoder, zhao2017pyramid, lin2017refinenet, zhang2018exfuse}. State of the art methods are based on Fully Convolutional Neural Networks \cite{long2015fully, badrinarayanan2017segnet} and use different %architectural and/or algorithmic 
strategies to condition pixel-level annotations on their global context, \eg using multiple scales \cite{zhao2017pyramid,lin2017refinenet,chen2017rethinking, chen2017deeplab, zhang2018exfuse, chen2018encoder} and/or modeling spatial dependencies \cite{chen2017rethinking,ghiasi2016laplacian}.  %Since the semantic information of near pixels is highly correlated, it is common to resort to multi-scale context information that can be extracted using different strategies. 
%For instance, different works exploit multi-scale information  and/or refinement strategies on the segmentation masks\massi{ADD CITs}.
The vast majority of semantic segmentation methods considers an offline setting, \ie they assume that training data for all classes is available beforehand. 
%Despite these progresses, very few works went beyond the offline scenario, \ie where all data of all classes are available beforehand. 
 %While progresses have been made on topics such few and zero shot learning \massi{CITS}, 
%and very few works discussed the problem of \icl\ in semantic segmentation 
To our knowledge, the problem of \icl\ in semantic segmentation has been addressed only in
\cite{ozdemir2018learn,ozdemir2019extending,tasar2019incremental,michieli2019incremental}. Ozdemir \etal \cite{ozdemir2018learn,ozdemir2019extending} describe an \icl\ approach for medical imaging, extending a standard image-level classification method \cite{li2017learning} to segmentation and devising a strategy to select relevant samples of old datasets for rehearsal. Taras 
\etal proposed a similar approach for segmenting remote sensing data. %However, their analysis is limited to medical images and no benchmarking is performed against standard \icl\ methods adapted to this task. %Moreover, they mention the fact that each learning step has a different background, but without tackling this problem.
Differently, Michieli \etal
\cite{michieli2019incremental} consider \icl\ for semantic segmentation 
in a particular setting where labels are provided for old classes while learning new ones. Moreover, they assume the novel classes to be never present as background in pixels of previous learning steps. These assumptions strongly limit the applicability of their method. % in real applications. %These assumptions are not realistic given (i) the cost of manually labeling all the classes for the new datasets and (ii) the impossibility of knowing future target classes beforehand without breaking the \icl\ assumptions.  %Moreover, as in \cite{ozdemir2018learn}, no benchmarking has been performed against standard \icl\ approaches.

Here we propose a more principled formulation of the \icl\ problem in semantic segmentation. %, without making any assumption or restriction on the classes contained in the images.
In contrast with previous works, we do not limit our analysis to medical \cite{ozdemir2018learn} or remote sensing data \cite{tasar2019incremental} and we do not impose any restrictions on how the label space should change across different learning steps \cite{michieli2019incremental}.  Moreover, 
%this work is 
we are the first to provide a comprehensive  experimental evaluation of \sota \icl\ methods on commonly used semantic segmentation benchmarks % scenario % semantic segmentation scenario. Finally, we are the first 
and to explicitly introduce and tackle the semantic shift of the background class, % in \icl,%the various learning steps, 
a problem recognized but largely overseen by previous works \cite{michieli2019incremental}.

\myparagraph{Incremental Learning.}
%A well known limitation of neural networks is catastrophic forgetting \cite{mccloskey1989catastrophic} \ie the tendency to erase past knowledge when we try to incrementally add new concepts to the network.% that happens when new concepts are added to the network. 
%How to incrementally add new classes to an already trained model 
The problem of catastrophic forgetting \cite{mccloskey1989catastrophic} has been extensively studied for image classification tasks \cite{de2019continual}. 
{Previous works can be grouped in three categories \cite{de2019continual}: replay-based \cite{rebuffi2017icarl, castro2018end, shin2017continual, hou2019learning, wu2018memory, ostapenko2019learning}, regularization-based \cite{kirkpatrick2017overcoming,chaudhry2018riemannian,zenke2017continual,li2017learning, dhar2019learning}, and parameter isolation-based \cite{mallya2018packnet, mallya2018piggyback, rusu2016progressive}.
In replay-based methods, examples of previous tasks are either stored \cite{rebuffi2017icarl, castro2018end, hou2019learning, wu2019large} or generated \cite{shin2017continual, wu2018memory, ostapenko2019learning} and then replayed while learning the new task. % to retain the old knowledge. % However, storing or generating samples is very costly and thus, we do not consider this approach in this work.
Parameter isolation-based methods \cite{mallya2018packnet, mallya2018piggyback, rusu2016progressive} assign a subset of the parameters to each task to prevent forgetting.} %In general, these methods are employed in task-incremental learning and are not suited to perform incremental class learning.
Regularization-based methods can be divided in prior-focused and data-focused. 
The former \cite{zenke2017continual, chaudhry2018riemannian, kirkpatrick2017overcoming, aljundi2018memory} define knowledge as the parameters value, constraining the learning of new tasks by penalizing changes of important parameters for old ones. 
%These methods mostly differ on how parameters relevance is computed.
The latter \cite{li2017learning, dhar2019learning} exploit distillation \cite{hinton2015distilling} and use the distance between the activations produced by the old network and the new one as a regularization term to prevent catastrophic
forgetting. %These methods mainly differ for the objective of the distillation. 

Despite these progresses, very few works have gone beyond image-level classification. A first work in this direction is \cite{shmelkov2017incremental} which considers \icl\ in object detection %. In \cite{shmelkov2017incremental}, the authors 
proposing a distillation-based method adapted from \cite{li2017learning} for tackling novel class recognition and bounding box proposals generation. 
%a method based on adapting distillation-based techniques \cite{li2017learning} for tackling novel class extension and proposals. 
In this work we also take a similar approach to \cite{shmelkov2017incremental} and we resort on distillation. However, here we propose to address the problem of modeling the background shift which is peculiar of the semantic segmentation setting. %show how the  context of semantic segmentation, the only works who tried to tackle \icl\ are \cite{ozdemir2018learn,michieli2019incremental}%Some recent works studied \icl\ in other tasks \cite{shmelkov2017incremental, ozdemir2018learn, michieli2019incremental}. 

\begin{figure*}
    \centering
    \includegraphics[width=0.95\textwidth]{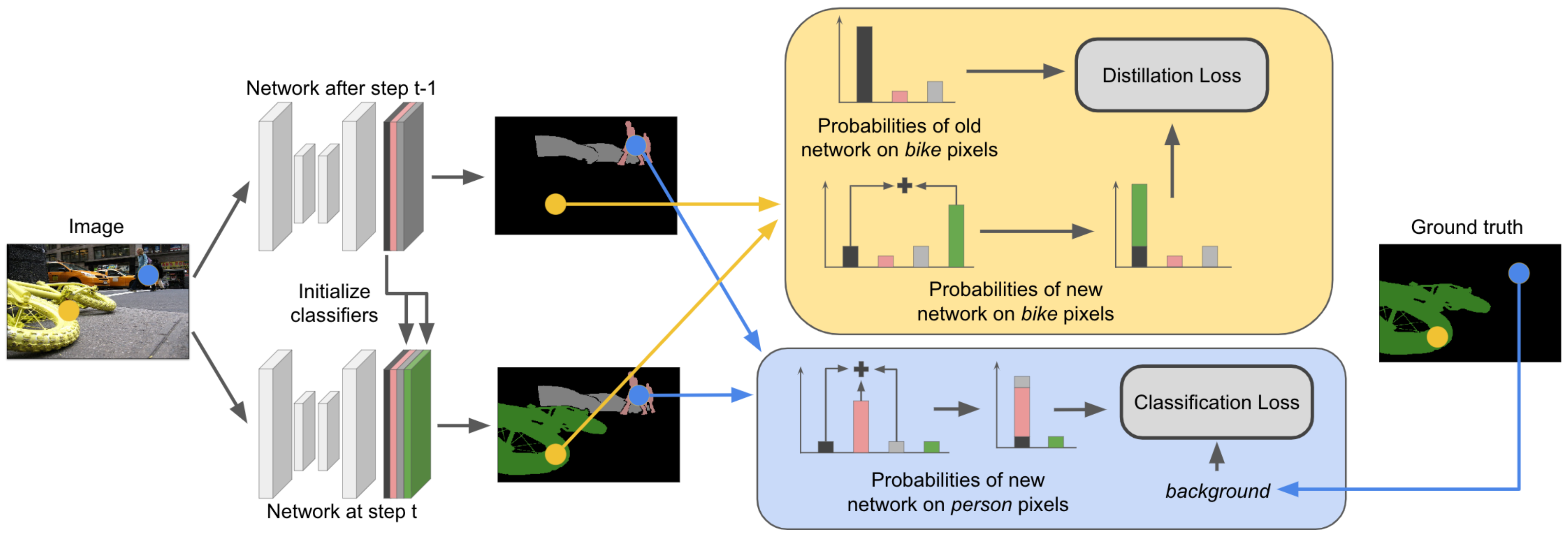}
    \vspace{-10pt}
    \caption{Overview of our 
    %the proposed \icl\ 
    method. 
    %for semantic segmentation. 
    At learning step $t$ an image is processed by the old (top) and current (bottom) models, mapping the image to their respective output spaces. As in standard \icl\ methods, we apply a cross-entropy loss to learn new classes (blue block) and a distillation loss to preserve old knowledge (yellow block). In this framework, we model the semantic changes of the background 
 across different learning steps by (i) initializing the new classifier using the weights of the old background one (left), (ii) comparing the pixel-level background ground truth in the cross-entropy with the probability of having either the background (black) or an old class (pink and grey bars) and (iii) relating the background probability given by the old model in the distillation loss with the probability of having either the background or a novel class (green bar).  \vspace{-10pt}}
    \label{fig:method}
\end{figure*}
 
\section{Method}
\label{sec:method}
%In the following, we first formalize the problem of \icl\ for semantic segmentation (Sec. \ref{sec:problem}). Then, in Sec. \ref{sec:our-method}, we formulate a general objective function for solving this task and we show how we can implement it in order to explicitly model background information, thus effectively addressing \icl\ in semantic segmentation.

\subsection{Problem Definition and Notation}
\label{sec:problem}
%In this work we tackle the problem of \icl\ for semantic segmentation. In particular, 
%Our goal is to progressively augment the output space of a pretrained segmentation model with novel classes for which no labeled data were available in the initial training phase. Furthermore, we assume that, while learning the novel categories, the initial training set is unavailable and no annotation is provided for original classes. This introduces an inherent bias of the learned model towards the novel classes, %which in turns causes 
%eventually leading to 
%the issue of 
%catastrophic forgetting.
%The goal of semantic image segmentation is to assign a semantic label to each pixel of an image. 

%\textbf{Incremental Learning for Semantic Segmentation}
Before delving into the details of \icl\ for semantic segmentation, we first introduce the task of semantic segmentation.
%Given a predetermined set of $C$ semantic
%classes encoded by $\mathcal{Y}= \{1,\dots, C\}$, 
Let us denote as $\mathcal{X}$ the input space (\ie the image space) and, without loss of generality, let us assume that each image $x\in\mathcal{X}$ is composed by a set of pixels $\set I$ with constant cardinality $|\set I|=N$. %Given an image $x$ and a label space $\set Y$, the goal of semantic segmentation is to assign each pixel $x_i\in x$ a label $y_i \in \mathcal{Y}$, representing its semantic class. Out-of-class pixels can be assigned a special class, \ie the background class $\mathtt{b}\in\set Y$. The output space is a segmentation mask $(\mathcal{Y})^{N}$, with the latter denoting the product set of $N$-tuples with elements in $\mathcal{Y}$. 
The output space is defined as $\mathcal{Y}^{N}$, with the latter denoting the product set of $N$-tuples with elements in a label space $\mathcal{Y}$. Given an image $x$ the goal of semantic segmentation is to assign each pixel $x_i$ of image $x$ a label $y_i \in \mathcal{Y}$, representing its semantic class. Out-of-class pixels can be assigned a special class, \ie the background class $\mathtt{b}\in\set Y$. 
% Given a training set $\mathcal{T} \subset \mathcal{X}\times (\mathcal{Y})^N$, the mapping is realized by learning a model $f_{\theta}$ with parameters $\theta$ mapping elements $x\in\mathcal{X}$ to a pixel-wise class probability vector $q_x$, with the final pixel-wise prediction obtained as $y^* = \{ \argmax_{c\in\mathcal{Y}} q_x(i,c)\}_{i=1}^{N}$.
% Given a training set $\mathcal{T} \subset \mathcal{X}\times (\mathcal{Y})^N$, the mapping is realized by learning a model $f_{\theta}$ with parameters $\theta$ from the image pixels space $\mathcal{X}$ to their corresponding segmentation mask $(\mathcal{Y})^{N}$, \ie $\Phi_{\theta} : \mathcal{X} \mapsto (\mathcal{Y})^{N}$. \fabio{In the following, given a pixel $i$, we consider $\Phi_{\theta}(x)[i] = \argmax_c f_{\theta}(x)[i,c]$, where $c\in \mathcal{Y}$ and $f_{\theta}$ is a neural network outputting a pixel-wise class probability.}
{Given a training set $\mathcal{T} \subset \mathcal{X}\times \mathcal{Y}^N$, the mapping is realized by learning a model $f_{\theta}$ with parameters $\theta$ from the image space $\mathcal{X}$ to a pixel-wise class probability vector, \ie $f_{\theta} : \mathcal{X} \mapsto \mathcal{\real}^{N \times |\mathcal{Y}|}$.
%Given an image $x$, 
The output segmentation mask is obtained as $y^* = \{ \argmax_{c\in\mathcal{Y}} f_{\theta}(x)[i,c]\}_{i=1}^{N}$, where $f_{\theta}(x)[i,c]$ is the probability for class $c$ in pixel $i$.}

In the \icl\ setting, training is realized over multiple phases, called \textit{learning steps}, and each step introduces novel categories to be learnt. In other terms, during the $t_{\text{th}}$ learning step, the previous label set $\set Y^{t-1}$ is expanded with a set of new {classes} $\set C^t$, yielding a new label set $\set Y^t=\set Y^{t-1}\cup\set C^{t}$.
%, each corresponding to expanding the set of possible object categories, \ie for each step $t$ the output space $\mathcal{Y}^{t}= \mathcal{Y}^{t-1} \cup \mathcal{C}^t$ where $\mathcal{C}^t$ is the set comprising to the novel categories.
%Let us call each expansion of the model with a new set of classes as \textit{learning step}, with the initial training phase denoted as \textit{step 0}. 
At learning step $t$ we are also provided with a training set $\mathcal{T}^t \subset \mathcal{X}\times (\set {C}^{t})^N$ that is used in conjunction to the previous model $f_{\theta^{t-1}}:\set X\mapsto \real^{N\times |\set Y^{t-1}|}$ to train an updated model $f_{\theta^{t}}:\set X\mapsto \real^{N\times |\set Y^{t}|}$.
%The starting point of a learning step $t$ is a set of parameters $\theta^{t-1}$ corresponding to the model $f_{\theta^{t-1}}: \mathcal{X} \mapsto \mathcal{Y}^{t-1}$, while the output is $\theta^t$ associated to $f_{\theta^t}$ and $\mathcal{Y}^t$. 
%, where %$\theta^{t-1}$ is the set of learned parameters, $\mathcal{X}$ is the image space and $\mathcal{Y}^{t-1}_0$ is the output space at the previous learning step. In this context, 
%$\mathcal{Y}^{t-1}$ is the set of possible pixel-wise labelling using all the known classes accumulated from learning step $0$ to $t-1$. In particular, if we denote as $\mathcal{C}^t$ the set of semantic classes relative to the learning step $t$, the set of all known classes at a learning step $t$ is $\mathcal{K}^t = \cup_{i=0}^{t}\mathcal{C}^i$.
%Notice that for two different learning steps we assume that the corresponding $\mathcal{C}^j$ %set of classes 
%share only one semantic category: the special void/background class $\mathtt{b}$, \ie $\mathcal{C}^j \cap \mathcal{C}^k = \mathtt{b}$ if $j\neq k$. 
As in standard \icl, in this paper we assume the sets of labels $\mathcal{C}^t$ that we obtain at the different learning steps to be disjoint, except for the special void/background class $\con b$. %\massi{Qui come potremmo stressare che l'assunzione è ragionevole?}

\subsection{Incremental Learning for Semantic Segmentation with Background Modeling}
\label{sec:our-method}
A naive approach to address the \icl\ problem consists in retraining the model $f_{\theta^t}$ on each set $\mathcal{T}^t$ sequentially. When the predictor $f_{\theta^t}$ is realized through a deep architecture, this corresponds to fine-tuning the network parameters on the training set $\mathcal{T}^t$ initialized with the parameters $\theta^{t-1}$ from the previous stage. This approach is simple, but it leads to catastrophic forgetting. Indeed, when training using $\mathcal{T}^t$ no samples from the previously seen object classes %in $\mathcal{C}^{t-1}$, $\mathcal{C}^{t-2}$, $\dots$ 
are provided. This biases the new predictor $f_{\theta^t}$ %obtained with \eqref{eq:obj-ft} 
towards the novel set of categories in $\mathcal{C}^t$ to the detriment of the classes from the previous sets. In the context of \icl\ for image-level classification, a standard way to address this issue is coupling the supervised loss on $\mathcal{T}^t$ with a regularization term, either taking into account the importance of each parameter for previous tasks \cite{kirkpatrick2017overcoming,shin2017continual}, or by distilling the knowledge using the predictions of the old model $f_{\theta^{t-1}}$ \cite{li2017learning,rebuffi2017icarl,castro2018end}. We take inspiration from the latter solution to initialize the overall objective function of our problem. In particular, we  minimize a loss function of the form:
\begin{equation}
   \label{eq:obj-general}
    \mathcal{L}(\theta^t)= \frac{1}{|\mathcal{T}^t| %N M 
    }\sum_{(x,y)\in\mathcal{T}^t}
    \left(\ell^{\theta^t}_{ce}(x,y) + \lambda \ell^{\theta^t}_{kd}(x) \right)
\end{equation}
where $\ell_{ce}$ is a standard supervised loss (\eg cross-entropy loss), $\ell_{kd}$ is the distillation loss and $\lambda>0$ is a \hyper balancing the importance of the two terms. 
%\subsection{Incremental Learning for Semantic Segmentation with Background Modeling}
%\label{sec:our-method}

As stated in Sec. \ref{sec:problem}, differently from standard \icl\ settings considered for image classification problems, in semantic segmentation we have that two different label sets $\mathcal{C}^s$ and $\mathcal{C}^u$  share the common void/background class $\mathtt{b}$. % $\mathcal{C}_i\cap\mathcal{C}_j=\mathtt{b}$ because in all sets there could be pixels without any known label and thus assigned to the background class. 
However, %it is easy to spot that 
 % it is reasonable to assume that 
{the distribution of the background class} changes across different incremental steps. In fact, background annotations given in $\set T^t$ refer to classes not present in $\set C^t$, that might belong to the set of seen classes $\set Y^{t-1}$ and/or to still unseen classes \ie $\set C^{u}$ with $u>t$  (see Fig. \ref{fig:teaser}). %  present \emph{either in past or future} sets \ie $\set C$ .In fact, background pixels in $\mathcal{Y}^{t-1}$ could have been assigned to a class in $\mathcal{C}^t$ and vice-versa (see Fig. \ref{fig:teaser}). 
In the following, we show how we account for the semantic shift in the distribution of the background class by revisiting standard choices for the general objective defined in Eq. \eqref{eq:obj-general}. %losses $\ell_{ce}$ and $\ell_{kd}$. % and effectively addressing the incremental class learning problem in semantic segmentation. %Finally, we will also describe a simple strategy for initializing, at each learning step $t$, the parameters of the classifier for the newly introduced classes, which leads to a more stable training and an actual performance boost. %modifying the supervised cross-entropy (Eq.\eqref{eq:CE}) and the distillation loss (Eq.\eqref{eq:std-distill}) and, thus, how we effectively tackle the incremental class learning problem in semantic segmentation. We also describe an effective strategy for initializing, at each learning step $t$, the parameters of the classifier for the newly introduced classes.
%To tackle this problem, different works in incremental learning for classification \cite{li2017learning,rebuffi2017icarl,castro2018end} proposed to add to the objective loss a regularization term based on the notion of distillation \cite{hinton2015distilling}.

\myparagraph{Revisiting Cross-Entropy Loss. } In Eq.\eqref{eq:obj-general}, a possible choice for $\ell_{ce}$ is the standard cross-entropy loss computed over all image pixels: 
\begin{equation}
   \label{eq:CE}
 %\ell^{\theta^t}_{ce}(x,y) = -\frac{1}{|\set I|}\sum_{i \in\set I}\sum_{c \in\set C^t} \mathbb{1}_{c=y(i)}\log q_x^t(i,c)
 \ell^{\theta^t}_{ce}(x,y) = -\frac{1}{|\set I|}\sum_{i \in\set I}\log q_x^t(i,y_i)\,,
\end{equation} 
where $y_i \in \set Y^t$ is the ground truth label associated to pixel $i$ and $q_x^t(i,c)=f_{\theta^t}(x)[i,c]$. %is the probability for class $c$ in pixel $i$ given by the model $f_{\theta^t}$. %, \ie $q_x^t(i,c)=f_{\theta^t}(x)[i,c]$, with $f_{\theta^t}(x)[i,c]$ denoting the probability of class $c$ in pixel $i$ as predicted by $f_{\theta^t}(x)$.  \massi{sistemare notazione f, specificando index e dimensioni}.

The problem with Eq.\eqref{eq:CE} is that the training set $\mathcal{T}^t$ we use to update the model %in each learning step $t$ 
only contains information about novel classes in $\mathcal{C}^t$. However, the background class in $\mathcal{T}^t$ might include also pixels associated to the previously seen classes in $\mathcal{Y}^{t-1}$. In this paper, we argue that, without explicitly taking into account this aspect, the catastrophic forgetting problem would be even more severe. In fact, we would drive our model to predict the background label $\mathtt{b}$ for pixels of old classes, further degrading the capability of the model to preserve semantic knowledge of past categories. To avoid this issue, in this paper we propose to modify the cross-entropy loss in Eq.\eqref{eq:CE} as follows: 
\begin{equation}
   \label{eq:our-CE}
    \ell^{\theta^t}_{ce}(x,y) = -\frac{1}{|\set I|}\sum_{i \in\set I}\log \tilde{q}_x^t(i,y_i)\,,
 \end{equation}
 where:
\begin{equation}
    \label{eq:cases-ce}
    \tilde{q}_x^t(i,c) = \begin{cases}
      {q}_x^t(i,c)\;\;& \text{if}\ c\neq\mathtt{b}\\
      \sum_{k\in\mathcal{{Y}}^{t-1}}{q}_x^t(i,k)\;\;& \text{if}\ c=\mathtt{b}\,.
    \end{cases}
\end{equation}

Our intuition is that by using Eq.\eqref{eq:our-CE} we can update the model to predict the new classes and, at the same time, account for the uncertainty over the actual content of the background class. In fact, in Eq.\eqref{eq:our-CE} the background class ground truth is not directly compared with its probabilities ${q}_x^t(i,\con b) $ obtained from the current model $f_{\theta^t}$, but with the probability of having \textit{either an old class or the background}, as predicted by $f_{\theta^t}$ (Eq.\eqref{eq:cases-ce}). {A schematic representation of this procedure is depicted in Fig.~\ref{fig:method} (blue block).}
% This solution allows to model the fact that a pixel labeled as background in $\mathcal{T}^t$ may actually belong to any of the classes not in $\mathcal{C}^t$ or the background itself: this produces a positive effect by both (i) avoiding a bias on $\mathtt{b}$ and (ii) mitigating the catastrophic forgetting problem.
%Notice that, without taking into account the uncertainty in $\mathtt{b}$, we would risk to have the probability of old classes harm the performances on already known classes. Indeed, if we apply \eqref{eq:CE} over a pixel labeled as background for $\mathcal{C}^t$ but actually containing a class in $\mathcal{K}_t$ we will make more severe the catastrophic forgetting problem, since feaaboutres of old classes will be associated with the background classifier. 
It is worth noting that the alternative of ignoring the background pixels within the cross-entropy loss is a sub-optimal solution. In fact, this would not allow to adapt the background classifier to its semantic shift and to exploit the information that new images might contain about old classes. %Employing \eqref{eq:our-CE} instead allows to regularizing the predictions of the network for the background, thus avoiding to (i) experience a severe forgetting and (ii) to have a fixed and sub-optimal background representation.

\myparagraph{Revisiting Distillation Loss.} In the context of incremental learning, distillation loss \cite{hinton2015distilling} is a common strategy to transfer knowledge from the old model $f_{\theta^{t-1}}$ into the new one, preventing catastrophic forgetting. %, by distilling the knowledge from the old model $f_{\theta^{t-1}}$ into the new one. 
Formally, a standard choice for the distillation loss $\ell_{kd}$ is:
\begin{equation}
    \label{eq:std-distill}
     \ell^{\theta^t}_{kd}(x,y) = -\frac{1}{|\set I|}\sum_{i \in\set I}\sum_{c \in\set Y^{t-1}} q_x^{t-1}(i,c)\log \hat{q}_x^{t}(i,c)\,,
\end{equation}
where $\hat{q}_x^{t}(i,c)$ is defined as the probability of class $c$ for pixel $i$ given by $f_{\theta^t}$ but re-normalized across all the classes in $\set Y^{t-1}$ \ie:
\begin{equation}
    \label{eq:cases-kd}
    \hat{q}_x^{t}(i,c)= \begin{cases}
       0 \;\;& \text{if}\ c\in \set C^t\setminus\{\mathtt{b}\}\\
      {q}_x^{t}(i,c)/\sum_{k\in \set Y^{t-1}} q_x^{t}(i,k) \;\;& \text{if}\ c\in \set Y^{t-1}\,.
    \end{cases}
\end{equation}
%where $q_{t-1}(x)=\sigma_{t-1} (f_{\theta^{t-1}}(x))$ and $q_t^\prime(x)$ are the output probabilities of the current model by applying the softmax operator only on old classes, \ie $q_t^\prime(x)=\sigma_{t-1} (f_{\theta^t}(x))$. %In \eqref{eq:obj-ft}, $\lambda$ is a trade-off parameter between the two losses. 

The rationale behind $\ell_{kd}$ is that $f_{\theta^t}$ should produce activations close to the ones produced by $f_{\theta^{t-1}}$. This regularizes the training procedure in such a way that the parameters $\theta^t$ are still anchored to the solution found for recognizing pixels of the previous classes, \ie $\theta^{t-1}$. %Notice indeed that the softmax function is here defined across the set of classes known before starting the training step $t$, ignoring all the novel classes $\mathcal{C}_t$ added in $f_{\theta^n}$. 

The loss defined in Eq.\eqref{eq:std-distill} has been used either in its base form or variants in different contexts, from incremental task \cite{li2017learning} and class learning \cite{rebuffi2017icarl,castro2018end} in object classification to complex scenarios such as detection \cite{shmelkov2017incremental} and segmentation \cite{michieli2019incremental}. Despite its success, it has a fundamental drawback in semantic segmentation: it completely ignores the fact that the background class is shared among different learning steps. % and that background annotations in $\set T^t$ refer to classes not present in $\set C^t$ but might refer to seen classes $\set Y^{t-1}$ or unseen ones in $\set C^{t+m}$ with $m>0$. 
While with Eq.\eqref{eq:our-CE} we tackled the first problem linked to the semantic shift of the background (\ie $\mathtt{b} \in \set T^{t}$ contains pixels of $\set Y^{t-1}$), we use the distillation loss to tackle the second: {annotations for background in $\set T^s$ with $s<t$ might include pixels of classes in $\set C^t$.}%$\mathtt{b}$ in $ \set T^{t}$ $f_{\theta^{t-1}}$ might contain classes in $\set Y^{t-1}$}.  % predicted by it might contain pixels of unseen classes for . %In the following section we describe how we incorporate the background in a principled way within the incremental learning steps.

%To account for the background in a principled way, we also propose to revise the distillation loss. % to this scenario means to explicitly taking into account the fact that our semantic knowledge is shifting after each learning step.
%In the considered setting, we should take into account the fact that at each learning step what was previously labeled as background may belong to the semantic class we are currently aiming to learn. 
{From the latter considerations, the background probabilities assigned to a pixel by the old predictor $f_{\theta^{t-1}}$ and by the current model $f_{\theta^{t}}$ do not share the same semantic content.} % of the background probability assigned by $f_{\theta^{t}}$. %, due to the inherent shift of the background class. 
More importantly, $f_{\theta^{t-1}}$ might predict as background pixels of classes in $\set C^t$ that we are currently trying to learn. Notice that this aspect is peculiar to the segmentation task and it is not considered in previous incremental learning models. However, in our setting we must explicitly take it into account to perform a correct distillation of the old model into the new one. To this extent we define our novel distillation loss by rewriting $\hat{q}_x^t(i,c)$ in Eq.\eqref{eq:cases-kd} as: % During an incremental learning learning the new classifier we must take into account the semantic shift of 
%Since the class background changes form incremental step to incremental step, we cannot perform simple pixel-wise distillation on it. We can thus either ignore it, which is a suboptimal solution, or comparing the probability of the old background with the probability of the new background and the new classes. We modify distillation in such way:
% \begin{equation}
%     \label{eq:our-distill}
%     \ell_{ukd}(x)=\sum_{k \in \mathcal{{Y}}^{t-1}} q_{t-1}(x)_k \log \hat{q}_t(x)_k 
% \end{equation}
% where:
\begin{equation}
    \label{eq:cases-ukd}
    \hat{q}_x^{t}(i,c)= \begin{cases}
      {q}_x^{t}(i,c)\;\;& \text{if}\ c\neq\mathtt{b}\\
      \sum_{k\in \set C^t}q_x^{t}(i,k)\;\;& \text{if}\ c=\mathtt{b}\,.
    \end{cases}
\end{equation}
Similarly to Eq.\eqref{eq:std-distill}, we still compare the probability of a pixel belonging to seen classes assigned by the old model, with its counterpart computed with the current parameters $\theta^t$. However, differently from classical distillation, in Eq.\eqref{eq:cases-ukd} the probabilities obtained with the current model are kept unaltered, \ie normalized across the whole label space $\set {Y}^t$ and not with respect to the subset $\set Y^{t-1}$ (Eq.\eqref{eq:cases-kd}). More importantly, the background class probability as given by $f_{\theta^{t-1}}$ is not directly compared with its counterpart in $f_{\theta^t}$, but with the probability of having \textit{either a new class or the background}, as predicted by $f_{\theta^t}$ ({see Fig. \ref{fig:method}, yellow block}).  %\massi{Figure \ref{fig:method} shows a schematic representation of this procedure. (yellow block).}

We highlight that, with respect to Eq.\eqref{eq:cases-kd} and other simple choices %(\eg excluding $\con b$ from Eq.\eqref{eq:cases-kd}, applying distillation only on pixels labeled as $\con b$)
(\eg excluding $\con b$ from Eq.\eqref{eq:cases-kd}) this solution has two advantages. First, we can still use the full output space of the old model to distill knowledge in the current one, without any constraint on pixels and classes. Second, we can propagate the uncertainty we have on the semantic content of the background in $f_{\theta^{t-1}}$ %of the old parameters regarding the actual semantic content of the background 
without penalizing the probabilities of new classes we are learning in the current step $t$. %Third we do not impose any constraint on which pixels it should be applied %TODO\textcolor{red}{Finally, we would like to remark that Eq.\eqref{eq:our-distill} is the counterpart of Eq.\eqref{eq:our-CE}, when the ground truth label $y$ is replaced by the old model predictions $q_{t-1}$, producing an elegant solution to the semantic shift of $\mathtt{b}$ and the catastrophic forgetting problem.}

%$\hat{q}_n^t(x)_k={q}_n^t(x)_k$ if $k \neq b$ but, in case of the background, it is defined as:
% \begin{equation}
%       \hat{p}_b^n(x)  = \sum_{c\in \{\mathcal{N},b\}}p_c^n(x)
% \end{equation}

\myparagraph{Classifiers' Parameters Initialization.} As discussed above, the background class $\con b$ is a special class devoted to collect the probability that a pixel belongs to an unknown object class. %when the current classification models is not able to assign a pixel to any known object class, it outputs the background label $\con b$. 
In practice, at each learning step $t$, the novel categories in $\set C^t$ are unknowns for the old classifier $f_{\theta^{t-1}}$. As a consequence, unless the appearance of a class in $\set C^t$ is very similar to one in $\set Y^{t-1}$, it is reasonable to assume that $f_{\theta^{t-1}}$ will likely assign pixels of $\set C^t$ to $\con b$. % the novel categories to the background. 
Taking into account this initial bias on the predictions of $f_{\theta^{t}}$ on pixels of $\set C^t$, it is detrimental to randomly initialize the classifiers for the novel classes. In fact a random initialization would provoke a misalignment among the features extracted by the model (aligned with the background classifier) and the random parameters of the classifier itself. Notice that this could lead to possible training instabilities while learning novel classes since the network could initially assign high probabilities for pixels in $\set C^t$ to $\con b$. % could lead to possible training instabilities. In fact, since the features extracte by the old model for the ntends to extract features  since the initial predictions for novel classes would be inherently unaligned with the features extracted by the old model, leading to possible training instabilities. %While this might not always happen (\eg the appearance of some images corresponding to class in $\mathcal{C}^t$ is very similar to those of one of the classes in $\mathcal{Y}^{t-1}$), \textcolor{red}{in our scenario is more probable since we assume the possibility of having pixels of classes in $\mathcal{C}^t$ labeled as background in the previous learning step. Taking into account this initial training bias for novel classes, it is detrimental to randomly initialize the classifiers for the novel classes since it would (i) not respect the priors given by the actual problem statement and (ii) lead to possible training instabilities}.  

% To address this issue, we propose to explicitly align the classifiers for the novel classes in $\set C^t$ to the background classifier of $f_{\theta^{t-1}}$. To this extent, we initialize the classifiers of the novel classes in such a way that given an image $x$ and a pixel $i$, the probability of the background $q_x^{t-1}(i,\con b)$ is uniformly spread among $\set C^{t}$, \ie $q_x^{t}(i,c)=q_x^{t-1}(i,\con b)/|\set C^t|$, where $|\mathcal{C}^t|$ is the number of new classes (notice that $\mathtt{b} \in \mathcal{C}^t$).

To address this issue, we propose to initialize the classifier's parameters for the novel classes in such a way that given an image $x$ and a pixel $i$, the probability of the background $q_x^{t-1}(i,\con b)$ is uniformly spread among the classes in $\set C^{t}$, \ie $q_x^{t}(i,c)=q_x^{t-1}(i,\con b)/|\set C^t|\; \forall c \in \set C^t$, where $|\mathcal{C}^t|$ is the number of new classes (notice that $\mathtt{b} \in \mathcal{C}^t$). 
To this extent, let us consider a standard fully connected classifier and let us denote as $\{\omega^{t}_c, \beta^{t}_c\}\in\theta^t$ the classifier parameters for a class $c$ at learning step $t$, with $\omega$ and $\beta$ denoting its weights and bias respectively. We can initialize $\{\omega^t_c, \beta^t_c\}$ as follows:
\begin{align}
    \label{eq:init-cases}
    \omega_c^{t}&=\begin{cases}
     \omega_{\con b}^{t-1} \;\;& \text{if}\ c \in \mathcal{C}^t\\
      \omega^{t-1}_c \;\;& \text{otherwise}\\
    \end{cases}\\
        \label{eq:init-cases2}
    \beta_c^{t}&=\begin{cases}
      \beta_{\con b}^{t-1} - \log(|\set C^t|)\;\;& \text{if}\ c \in \mathcal{C}^t\\
      \beta^{t-1}_c \;\;& \text{otherwise}\\
    \end{cases}
\end{align}
% \begin{equation}
%     \label{eq:init-cases}
%     \omega_c^{t}=\begin{cases}
%       \omega^{t-1}_c \;\;\text{and}\;\; \beta_c^{t}=\beta_c^{t},\;\;\;\;\;\;\;\;\;\;\;\; \text{if}\ c \in \mathcal{Y}^{t-1}\wedge c\neq\mathtt{b}\\
%       \omega_c^{t}=\omega_\mathtt{b}^{t-1} \;\;\text{and}\;\; \beta_{t}^c=\beta_\mathtt{b}^{t-1} - \log(|\mathcal{C}^t|),\;\;\;\;\;\;\;\; \text{if}\ c \in \mathcal{C}^t\\
%     \end{cases}
% \end{equation}
 where $\{\omega_\mathtt{b}^{t-1},\beta_\mathtt{b}^{t-1} \}$ are the weights and bias of the background classifier at the previous learning step. The fact that the initialization defined in Eq.\eqref{eq:init-cases} and \eqref{eq:init-cases2} leads to $q_x^{t}(i,c)=q_x^{t-1}(i,\con b)/|\set C^t|\; \forall c \in \set C^t$ is easy to obtain from $q_x^{t}(i,c)\propto \exp(\omega_\mathtt{b}^{t}\cdot x + \beta^{t}_\mathtt{b})$. %For the full derivation please see the supplementary material.% simply derived by just uniformly dividing the probability of $\mathtt{b}$ from the old classifier among the new classes and $\mathtt{b}$, \ie: 
% \begin{equation}
%       \label{eq:init-derivation}
%       \log{\frac{q_x^{t-1}(i,\con b)}{|\set C^t|}} \propto \frac{e^{\omega_\mathtt{b}^{t-1}\cdot x + \beta^{t-1}_\mathtt{b}}}{|\mathcal{C}^t|}=\text{exp}\{{\omega_\mathtt{b}^{t-1}\cdot x + \beta^{t-1}_\mathtt{b}-\log(|\mathcal{C}^t|)}\}
% \end{equation}
% for $c \in \mathcal{{C}}^t$.

As we will show in the experimental analysis, this simple initialization procedure brings benefits in terms of both improving the learning stability of the model and the final results, since it eases the role of the supervision imposed by Eq.\eqref{eq:our-CE} while learning new classes and follows the same principles used to derive our distillation loss (Eq.\eqref{eq:cases-ukd}).% \massi{Possiamo aggiungere qualcosa per rinforzare la scelta?}

%\massi{Full pipeline. The full pipeline is shown in Fig.\ref{fig:method}.}

% \textbf{Implementation details}: For practical reasons, we can rewrite the loss relative to the background with two \textit{logsumofexps}:
% \begin{align}
%     L_{ukd}(x,b)&= p_b^o(x) \log \sum_{i\in \{\mathcal{N},b\}}e^{\theta^i^{n\;\intercal} x + \beta^n_i}\\ &- p_b^o(x) \log \sum_{j\in \{\mathcal{M},\mathcal{N},b\}}e^{\theta^j^{n\;\intercal} x + \beta^n_j}
% \end{align}

% SINGLE
\begin{table*}[t]
\centering
\setlength{\tabcolsep}{3pt} % Default value: 6pt
\small
\caption{Mean IoU on the Pascal-VOC 2012 dataset for different incremental class learning scenarios.\vspace{-8pt}}
\label{tab:pascal}
\begin{tabular}{l||cc|c||cc|c||cc|c||cc|c||cc|c||cc|c}
\multicolumn{1}{c}{}    & \multicolumn{6}{c}{\textbf{{19-1}}}   & \multicolumn{6}{c}{{\textbf{15-5}}} & \multicolumn{6}{c}{{\textbf{15-1}}}    \\
\multicolumn{1}{c||}{}    & \multicolumn{3}{c||}{\bf{Disjoint}}        & \multicolumn{3}{c||}{\bf{Overlapped}}  & \multicolumn{3}{c||}{\bf{Disjoint}}     & \multicolumn{3}{c||}{\bf{Overlapped}}  & \multicolumn{3}{c||}{\textbf{Disjoint}}      & \multicolumn{3}{c}{\textbf{Overlapped}} \\
\bf{Method} & \it{1-19}  & \it{20}   & \it{all}  & \it{1-19}  & \it{20}   & \it{all}     & \it{1-15}  & \it{16-20}   & \it{all}   & \it{1-15}  & \it{16-20}   & \it{all} & \it{1-15}  & \it{16-20}   & \it{all}  & \it{1-15}  & \it{16-20}   & \it{all}     \\ \hline
{FT }     & 5.8     & 12.3    & 6.2     & 6.8       & 12.9      & 7.1      & 1.1    & 33.6    & 9.2      & 2.1     & 33.1  & 9.8    & 0.2        & 1.8        & 0.6        & 0.2        & 1.8        & 0.6       \\
{PI \cite{zenke2017continual}}     & 5.4     & 14.1    & 5.9     & 7.5       & 14.0      & 7.8      & 1.3    & 34.1    & 9.5      & 1.6     & 33.3  & 9.5    & 0.0 &	1.8 &	0.4   & 0.0        & 1.8        & 0.5 \\
{EWC \cite{kirkpatrick2017overcoming}}    & 23.2    & 16.0    & 22.9    & 26.9      & 14.0      & 26.3     & 26.7   & 37.7    & 29.4     & 24.3    & 35.5  & 27.1   & 0.3        & 4.3        & 1.3        & 0.3        & 4.3        & 1.3       \\
{RW \cite{chaudhry2018riemannian}}     & 19.4    & 15.7    & 19.2    & 23.3      & 14.2      & 22.9     & 17.9   & 36.9    & 22.7     & 16.6    & 34.9  & 21.2   & 0.2        & 5.4        & 1.5          & 0.0        & 5.2        & 1.3 \\
{LwF \cite{li2017learning} }    & 53.0    & 9.1     & 50.8    & 51.2      & 8.5       & 49.1     & 58.4   & 37.4    & 53.1     & 58.9    & 36.6  & 53.3   & 0.8        & 3.6        & 1.5        & 1.0        & 3.9        & 1.8       \\
{LwF-MC \cite{rebuffi2017icarl}} & 63.0    & 13.2    & 60.5    & 64.4      & 13.3      & 61.9     & 67.2   & 41.2    & 60.7     & 58.1    & 35.0  & 52.3   & 4.5        & 7.0        & 5.2    & 6.4        & 8.4        & 6.9 \\
{ILT \cite{michieli2019incremental}}    & 69.1    & 16.4    & 66.4    & 67.1      & 12.3      & 64.4     & 63.2   & 39.5    & 57.3     & 66.3    & 40.6  & 59.9   & 3.7        & 5.7        & 4.2        & 4.9        & 7.8        & 5.7       \\
\ours   & \bf{69.6} & \bf{25.6}   & \bf{67.4}      & \bf{70.2}    & \bf{22.1}       & \bf{67.8}    & \bf{71.8}     & \bf{43.3}    & \bf{64.7}    & \bf{75.5}     & \bf{49.4}  & \bf{69.0}  & \bf{46.2}       & \bf{12.9}       & \bf{37.9}       & \bf{35.1}       & \textbf{13.5}       & \textbf{29.7}             \\ \hline
{Joint} & 77.4&	78.0&	77.4&	77.4&	78.0&	77.4&	79.1&	72.6&	77.4&	79.1&	72.6&	77.4 & 79.1       & 72.6       & 77.4       & 79.1       & 72.6       & 77.4
\end{tabular}
\vspace{-10pt}
\end{table*}
\section{Experiments}
\label{sec:exp}
%In the following, we assess the performance of our method comparing it with \sota \icl\ approaches. Note that these methods were originally designed for classification tasks, thus segmentation is treated as a pixel-level classification problem. We perform our evaluation on two common datasets, Pascal-VOC 2012 \cite{pascal-voc-2012} and ADE20K \cite{zhou2017scene}, considering different numbers of incremental steps and classes. All the results are reported as mean Intersection-over-Union (mIoU) in percentage, averaged over all the classes of a learning steps and all the steps. %or provided on a per learning step basis. %The baselines incremental learning methods will be introduced in Sec. \ref{sec:baselines}, then we describe implementation details in Sec. \ref{sec:impdetails}, and we report results in Sec. \ref{sec:pascal}, \ref{sec:ade}.
\subsection{\icl\
 Baselines} \label{sec:baselines}
\vspace{-5pt}
We compare our method against standard \icl\ baselines, 
 originally designed for classification tasks,
 on the considered segmentation task, 
 thus segmentation is treated as a pixel-level classification problem.
{Specifically, we report the results of six different regularization-based methods, three prior-focused and three data-focused approaches.} 
%Among the baselines we follow previous works on object detection \cite{shmelkov2017incremental} and we do not employ any method using rehearsal (\eg \cite{rebuffi2017icarl}) since they violate the standard \icl\ assumption regarding the unavailability of old data. %and (ii) any method can take advantage from using stored samples.

In the first category, we chose Elastic Weight Consolidation (EWC) \cite{kirkpatrick2017overcoming}, Path Integral (PI) \cite{zenke2017continual}, and Riemannian Walks (RW) \cite{chaudhry2018riemannian}. They employ different strategies to compute the importance of each parameter for old classes: EWC uses the empirical Fisher matrix, PI uses the learning trajectory, while RW combines EWC and PI in a unique model. We choose EWC since it is a standard baseline employed also in \cite{shmelkov2017incremental} and PI and RW since they are two simple applications of the same principle. Since these methods act at the parameter level, to adapt them to the segmentation task we keep the loss in the output space unaltered (\ie standard cross-entropy across the whole segmentation mask), computing the parameters' importance by considering their effect on learning old classes.

For the data-focused methods, we chose Learning without forgetting (LwF) \cite{li2017learning}, LwF multi-class (LwF-MC) \cite{rebuffi2017icarl} and the segmentation method of \cite{michieli2019incremental} (ILT).
We denote as LwF the original distillation based objective as implemented in Eq.\eqref{eq:obj-general} with basic cross-entropy and distillation losses, which is the same as \cite{li2017learning} except that distillation and cross-entropy share the same label space and classifier. LwF-MC is the single-head version of \cite{li2017learning} as adapted from \cite{rebuffi2017icarl}. It is based on multiple binary classifiers, with the target labels defined using the ground truth for novel classes (\ie $\set C^t$) and the probabilities given by the old model for the old ones (\ie $\set Y^{t-1}$). Since the background class is both in $\set C^t$ and $\set Y^{t-1}$ %we can choose if use the ground truth from $\set T^t$ or the probabilities given by $f_{\theta^{t-1}}$ as supervised signal for $\con b$. Since $\con b$ is both in $\set C^t$ and $\set Y^{t-1}$, 
we implement LwF-MC by a weighted combination of two binary cross-entropy losses, on both the ground truth and the probabilities given by $f_{\theta^{t-1}}$. %\massi{Since in this scenario using $\con b$ as the ground truth signal largely deteriorates the performances on old classes, while using $f_{\theta^{t-1}}$ as supervision prevents learning of novel ones (see supplementary material), we implement this strategy as a weighted combination among two binary cross-entropy losses, on both the ground truth and the probabilities of $f_{\theta^{t-1}}$. COSA FACCIAMO?}% it as supervised signal for $\con b$. APPENA ABBIAMO RES DI VOC POSSIAMO CAMBIARE} 
%To adapt LwF and LwF-MC to segmentation, we simply apply their losses at pixel level.
Finally, ILT \cite{michieli2019incremental} is the only method specifically proposed for \icl\ in semantic segmentation. It uses a distillation loss in the output space, as in our adapted version of LwF \cite{li2017learning} and/or another distillation loss in the features space, attached to the output of the network decoder. Here, we use the variant where both losses are employed.
As done by \cite{shmelkov2017incremental}, we do not compare with replay-based methods (\eg \cite{rebuffi2017icarl}) since they violate the standard \icl\ assumption regarding the unavailability of old data.

In all tables we report other two baselines: simple fine-tuning (FT) on each $\set T^t$ (\eg Eq.\eqref{eq:CE}) and training on all classes offline (Joint). The latter can be regarded as an upper bound. In the tables we denote our method as \ours (\expandednick). All results are reported as mean Intersection-over-Union (mIoU) in percentage, averaged over all the classes of a learning step and all the steps.

\subsection{Implementation Details}\label{sec:impdetails}
% network architecture
For all methods we use the Deeplab-v3 architecture \cite{chen2017rethinking} with a ResNet-101 \cite{he2016deep} backbone and output stride of 16. Since memory requirements are an important issue in semantic segmentation, we use in-place activated batch normalization, as proposed in \cite{rota2018place}. The backbone has been initialized using the ImageNet pretrained model \cite{rota2018place}. % learning hyper-parameters
We follow \cite{chen2017rethinking}, training the network with SGD and the same learning rate policy, momentum and weight decay. %caling the initial learning rate %we employ a learning rate policy where the initial learning rate is scaled 
%by $ (1- \frac{iter}{max\_iter})^{power}$ with $power = 0.9$. 
We use an initial learning rate of $10^{-2}$ for the first learning step and $10^{-3}$ for the followings, as in \cite{shmelkov2017incremental}.
We train the model with a batch-size of 24 for 30 epochs for Pascal-VOC 2012 and 60 epochs for ADE20K in every learning step. %Momentum and weight-decay are always set to 0.9 and $10^{-4}$ respectively.
% data preprocessing
We apply the same data augmentation of \cite{chen2017rethinking} %(\ie random scale and flipping) %to Pascal-VOC 2012 and ADE20K. We random scale the input images from 0.5 to 2.0 and we randomly flip it horizontally during training. Following \cite{chen2017rethinking}, 
and we crop the images to $512\times 512$ during both training and test.
For setting the \hypers of each method, we use the protocol of incremental learning defined in \cite{de2019continual}, using 20\% of the training set as validation.
The final results are reported on the standard validation set of the datasets.

%Since the used datasets do not release annotations on the test set, 
%The final results are reported on the full validation sets provided%we report the results on the validation set. Thus, we perform the validation protocol on a custom validation set obtained by splitting randomly in 80/20 the training set.
\begin{table*}[t]
\centering
\small
\setlength{\tabcolsep}{3pt} % Default value: 6pt
\caption{Mean IoU on the ADE20K dataset for different incremental class learning scenarios.\vspace{-8pt}}
\label{tab:ade}
\begin{tabular}{l||cc|c||cccccc|c||ccc|c}
\multicolumn{1}{c}{} & \multicolumn{3}{c}{{\textbf{100-50}}} & \multicolumn{7}{c}{{\textbf{100-10}}} & \multicolumn{4}{c}{{\textbf{50-50}}} \\
\textbf{Method}       & \textit{1-100} & \textit{101-150} & \textit{all}  & \textit{1-100} & \textit{100-110} & \textit{110-120} & \textit{120-130} & \textit{130-140} & \textit{140-150} & \textit{all}  & \textit{1-50} & \textit{51-100} & \textit{101-150} & \textit{all}  \\ \hline
FT     & 0.0   & 24.9    & 8.3  & 0.0   & 0.0    & 0.0    & 0.0    & 0.0    & 16.6   & 1.1  & 0.0  & 0.0    & 22.0    & 7.3  \\
LwF \cite{li2017learning}   & 21.1  & 25.6    & 22.6 & 0.1   & 0.0    & 0.4    & 2.6    & 4.6    & 16.9   & 1.7  & 5.7  & 12.9   & 22.8    & 13.9 \\
LwF-MC \cite{rebuffi2017icarl} & 34.2  & 10.5    & 26.3 & 18.7  &	2.5 &	8.7 &	4.1 &	6.5 &	5.1 &	14.3  & 27.8 & 7.0    & 10.4    & 15.1 \\
ILT \cite{michieli2019incremental}  & 22.9  & 18.9    & 21.6 & 0.3   & 0.0    & 1.0    & 2.1    & 4.6    & 10.7   & 1.4  & 8.4  & 9.7    & 14.3    & 10.8 \\
MiB   & \textbf{37.9}  & \textbf{27.9}    & \textbf{34.6} & \textbf{31.8}  & \textbf{10.4}   & \textbf{14.8}   & \textbf{12.8}   & \textbf{13.6}   & \textbf{18.7}   & \textbf{25.9} & \textbf{35.5} & \textbf{22.2}   & \textbf{23.6}    & \textbf{27.0} \\ \hline
Joint  & 44.3  & 28.2    & 38.9 & 44.3  & 26.1   & 42.8   & 26.7   & 28.1   & 17.3   & 38.9 & 51.1 & 38.3   & 28.2    & 38.9
\end{tabular}
\vspace{-10pt}
\end{table*}

\subsection{Pascal-VOC 2012} \label{sec:pascal}
PASCAL-VOC 2012 \cite{pascal-voc-2012} is a widely used benchmark that includes 20 foreground object classes. % and one background class. 
% It consists of 10582 images in the training set and 1449 images in the validation. %We report the results on the validation set. \massi{possiamo citare?} %Since the test set has not been released, we report the results on the validation set as it is the common protocol in literature \fabio{add citation}.
% How we make it incremental
Following \cite{michieli2019incremental,shmelkov2017incremental}, we define two experimental settings, depending on how we sample images to build the incremental datasets.
%Adopting the sampling strategy of 
Following \cite{michieli2019incremental}, we define an experimental protocol called the \textit{disjoint} setup: each learning step contains a unique set of images, whose pixels belong to classes seen either in the current or in the previous learning steps. Differently from \cite{michieli2019incremental}, at each step we assume to have only labels for pixels of novel classes, while the old ones are labeled as background in the ground truth. 
The second setup, that we denote as \textit{overlapped}, follows what done in \cite{shmelkov2017incremental} for detection: each training step contains all the images that have at least one pixel of a novel class, with only the latter annotated. It is important to note a difference with respect to the previous setup: images may now contain pixels of classes that we will learn in the future, but labeled as background. This is a more realistic setup since it does not make any assumption on the objects present in the images.

As done by previous works \cite{shmelkov2017incremental, michieli2019incremental}, we perform three different experiments concerning the addition of one class (\textit{19-1}), five classes all at once (\textit{15-5}), and five classes sequentially (\textit{15-1}), following the alphabetical order of the classes to split the content of each learning step. %Results in term of mean IoU are reported in Tab. \ref{tab:pascal}.

\myparagraph{Addition of one class \textit{(19-1)}.}
% 19-1
In this experiment, we perform two learning steps: the first in which we observe the first 19 classes, and the second where we learn the \textit{tv-monitor} class.
Results are reported in Table \ref{tab:pascal}. Without employing any regularization strategy, the performance on past classes drops significantly. FT, in fact, performs poorly, completely forgetting the first 19 classes. % especially on the first 19 classes.
Unexpectedly, using PI as a regularization strategy does not provide benefits, while EWC and RW improve performance of nearly 15\%. However, prior-focused strategies are not competitive with data-focused ones. In fact, LwF, LwF-MC, and ILT, outperform them by a large margin, confirming the effectiveness of this approch on preventing catastrophic forgetting. While ILT surpasses standard \icl\ baselines, our model is able to obtain a further boost. This improvement is remarkable for new classes, where we gain $11\%$ in mIoU, while do not experience forgetting on old classes. It is especially interesting to compare our method with the baseline LwF which uses the same principles of ours but without modeling the background. Compared to LwF we achieve an average improvement of about $15\%$, thus demonstrating the importance of modeling the background in \icl\ for semantic segmentation. These results are consistent in both the \textit{disjoint} and \textit{overlapped} scenarios.%in the new clas% . These methods obtain a large improvement especially on the first 19 classes, pointing out the strength of distillation-based approaches.  
% Our method outperforms all others
%Employing our proposed method, we further improve previous results, obtaining better accuracy especially on the new class, while the performance on previous classes is similar with distillation approaches.

% 15-5
\myparagraph{Single-step addition of five classes (\textit{15-5}).}
In this setting we add, after the first training set, the following classes: \textit{plant, sheep, sofa, train, tv-monitor}. Results are reported in Table \ref{tab:pascal}. 
% Comments
 %Overall, this scenario is easier than the \textit{19-1}, since almost the methods are able to obtain better performances, especially while learning new 5 classes.
Overall, the behavior on the first 15 classes is consistent with the 19-1 setting: FT and PI suffer a large performance drop, data-focused strategies (LwF, LwF-MC, ILT) outperform EWC and RW by far, while our method gets the best results, obtaining performances closer to the joint training upper bound. % on those classes close to the upper bound (joint training).
For what concerns the \textit{disjoint} scenario, our method improves over the best baseline of $4.6\%$ on old classes, of $2\%$ on novel ones and of $4\%$ in all classes. These gaps increase in the \textit{overlapped} setting where our method surpasses the baselines by nearly $10\%$ in all cases, clearly demonstrating its ability to take advantage of the information contained in the background class. %In this setting, our method largely improves the performances on both old and new classes.   %exceeding the  
%surpassing the baseline by nearly 10\% on the first 15 classes, and 3.8\% on the disjoint and 8.8\% on the overlapped setups on the new classes.

\myparagraph{Multi-step addition of five classes (\textit{15-1}).}
%In this setting we perform we perform an initial training step considering the first 15 classes and then we add one by one the remaining five classes.
% comments
This setting is similar to the previous one except that the last 5 classes are learned sequentially, one by one.
From Table \ref{tab:pascal} we can observe that performing multiple steps is challenging and existing methods work poorly for this setting, reaching performance inferior to 7\% on both old and new classes.
In particular, FT and prior-focused methods are unable to prevent forgetting, biasing their prediction completely towards new classes and demonstrating performances close to 0\% on the first 15 classes.
Even data-focused methods suffer a dramatic loss in performances in this setting, decreasing their score from the single to the multi-step scenarios of more than 50\% on all classes. %LwF also obtains very poor performance on those classes (0.8\% on disjoint and 1\% on overlapped setup). LwF-MC and ILT are the best performing methods among the prior works, but they are still far from an acceptable result. 
On the other side, our method is still able to achieve good performances. %, suffering a less dramatic drop which mostly affects new classes. 
Compared to the other approaches, \ours outperforms all baselines by a large margin in both old ($46.2\%$ on the \textit{disjoint} and $35.1\%$ on the \textit{overlapped}), and new (nearly $13\%$ on both setups) classes. As the overall performance drop ($11\%$ on all classes) shows, the \textit{overlapped} scenario is the most challenging one since it does not impose any constraint on which classes are present in the background. %The gap with LwF in this case is remarkable, since our method  especially thanks to its ability to reduce forgetting that allows to maintain decent performance on the first 15 classes. However, it is also able to learn better the new classes, obtaining nearly  mIoU on both data setups.
\begin{table}[t]
\small
\centering
\setlength{\tabcolsep}{2pt} % Default value: 6pt
\caption{Ablation study of the proposed method on the Pascal-VOC 2012 \textit{overlapped} setup. \textit{CE} and \textit{KD} denote our cross-entropy and distillation losses, while \textit{init} our initialization strategy. \vspace{-5pt}}% \massi{che nomi possiamo dargli?}} 
\label{tab:ablation}
\begin{tabular}{l||cc|c||cc|c||cc|c}
      \multicolumn{1}{c}{}& \multicolumn{3}{c}{\textbf{19-1}}          & \multicolumn{3}{c}{\textbf{15-5}}          & \multicolumn{3}{c}{\textbf{15-1}} \\
  &\it{1-19} & \it{20} & \it{all} & \it{1-15} & \it{16-20} & \it{all} & \it{1-15}  & \it{16-20}     & \it{all}     \\ \hline
LwF \cite{li2017learning}         & 51.2      & 8.5       & 49.1       & 58.9      & 36.6       & 53.3       & 1.0         & 3.9   & 1.8   \\
     + \textit{CE} & 57.6      & 9.9       & 55.2       & 63.2       & 38.1       & 57.0       & 12.0        & 3.7   & 9.9   \\
%+ KD & 63.0      & 11.8      & 60.5       & 74.3       & \textbf{51.5}       & 68.6       & 5.2         & 8.0   & 5.9   \\
+ \textit{KD}&   66.0      & 11.9      & 63.3       & 72.9       & 46.3       & 66.3       & 34.8        & 4.5   & 27.2  \\
+ \textit{init} & \textbf{70.2}      & \textbf{22.1}      & \textbf{67.8}      & \textbf{75.5}      & \textbf{49.4}      & \textbf{69.0}      & \textbf{35.1}       & \textbf{13.5} & \textbf{29.7}
\end{tabular}
\vspace{-13pt}
\end{table}

\myparagraph{Ablation Study.}
In Table \ref{tab:ablation} we report a detailed analysis of our contributions, considering the \textit{overlapped} setup. %In the table, \textit{CE} and \textit{KD} denote our cross-entropy and distillation losses respectively, while \textit{In} our initialization strategy. 
We start from the baseline LwF \cite{li2017learning} which employs standard cross-entropy and distillation losses.
%We started by comparing our proposed distillation (\ours KD) with the distillation used by LwF. It achieves better results on both previous and current classes, on all settings. 
We first add to the baseline our modified cross-entropy (\textit{CE}): this increases the ability to preserve old knowledge in all settings without harming (\textit{15-1}) or even improving (\textit{19-1}, \textit{15-5}) performances on the new classes. % This confirms the ability that our \textit{CE} has in preserving old knowledge, tackling the uncertainty on the background.
%\fabio{We then evaluate the use of our distillation loss (\textit{KD}) alone. Our \textit{KD} formulation allows to largely improve the performances of the baseline, in both old and new classes. With respect to adding only our \textit{CE}, the improvement on the novel classes is consistent across all settings while performances on the old ones are damaged only in the \textit{15-1} scenario. This confirms the fact that our \textit{KD} does not damage the learning process on novel classes correctly modeling the semantic shift of $\con b$ in successive learning step.}
Second, we add our distillation loss (\textit{KD}) to the model. 
%Combining both losses instead of using just one of them tends to 
Our \textit{KD} provides a boost on the performances for both old and new classes. The improvement on old classes is remarkable, especially in the 15-1 scenario (\ie 22.8\%). For the novel classes, the improvement is constant and is especially pronounced in the 15-5 scenario (7\%). Notice that this aspect is peculiar of our \textit{KD} since standard formulation work only on preserving old knowledge. %the distillation loss enforces knowledge preservation.  taking into account the possibility that novel classes where in the background of previous learning steps allows to not penaliz while . 
%For the novel classes however, this brings to performances which are in the middle of the ones achieved by adding only one of the loss. In the 15-1 scenario, the improvement over using just the \textit{CE} loss is remarkable (\ie 22.8\%) while adding only the \textit{KD} to the baseline brought to a lower improvement (4\%). 
This shows that the two losses provide mutual benefits. % and that i{considering only one direction of the semantic shift of the background (\ie from old classes to the current ground truth as in \textit{CE}) does not guarantee to achieve good performances.}
%Then, we evaluate the addition of our proposed cross entropy (\ours CE), combining it with either our distillation loss or the LwF loss. Adding it to LwF (LwF + \ours CE) provides a significant improvement, in particular on the \textit{15-1} setting where it outperforms all previous methods on old classes. However, the combination of our proposed losses (\ours no init) exceeds it on every setting, especially providing a large performance boost on past classes on the \textit{15-1} setting: $34.9\%$ vs $12.0\%$.
Finally, we add our classifiers' initialization strategy (\textit{init}). This component provides an improvement in every setting, especially on novel classes: it doubles the performance on the \textit{19-1} setting ($22.1\%$ vs $11.9\%$) and triplicates on the \textit{15-1} ($4.5\%$ vs $13.5\%$). This confirms the importance of accounting for the background shift at the initialization stage to facilitate the learning of new classes. %Compared to KD + \ours CE model, it provides an improvement on every setting, especially on new classes: it doubles the performance on the \textit{19-1} setting ($22.1\%$ vs $11.9\%$) and triplicates on the \textit{15-1} ($4.5\%$ vs $13.5\%$). %Remarkably, on the \textit{15-5} setting, using only our distillation (KD) achieves the best result on new classes, exceeding also our full method, but it forgets more about the first classes, achieving a lower overall result. \fabio{overall? distillation prevents forgetting, ce prevents bias toward background, init helps learning new classes} 

% qualitative results
\begin{figure*}[t]
     \centering
     \begin{subfigure}[b]{0.96\textwidth}
         \centering
         \includegraphics[width=\textwidth]{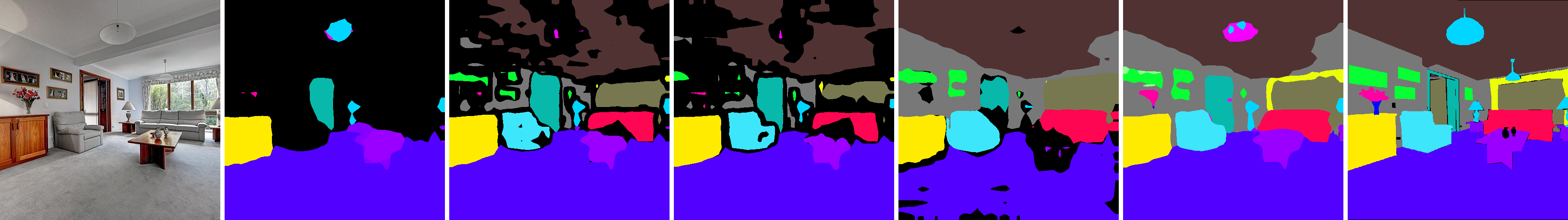}
     \end{subfigure}
     \hfill
     \begin{subfigure}[b]{0.96\textwidth}
         \centering
         \includegraphics[width=\textwidth]{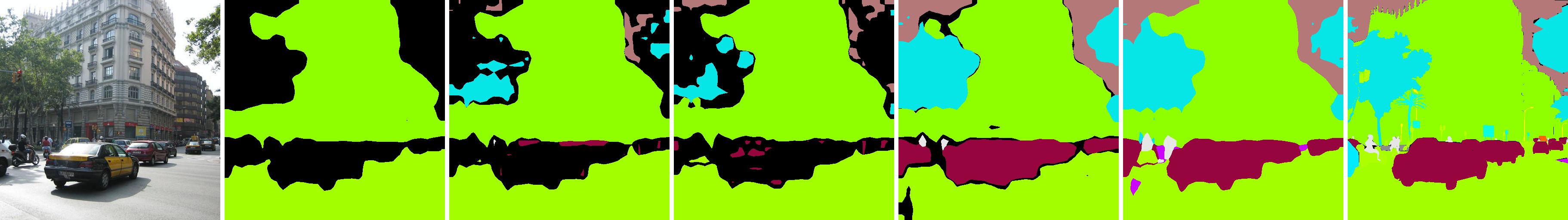}
     \end{subfigure}
     \hfill
     \begin{subfigure}[b]{0.96\textwidth}
         \centering
         \includegraphics[width=\textwidth]{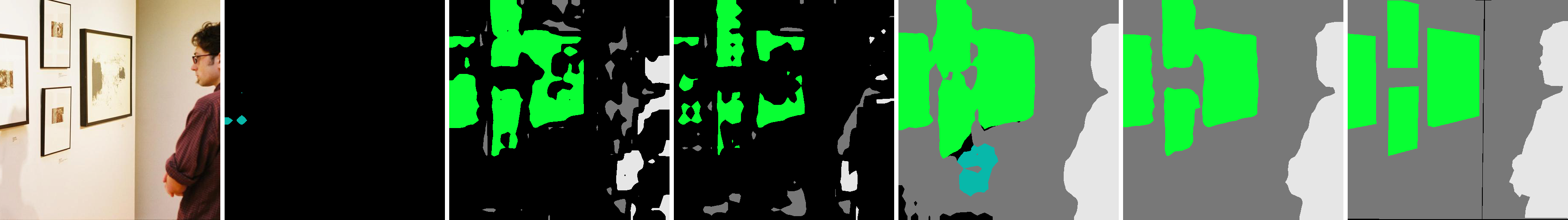}
     \end{subfigure}
          \hfill
     \begin{subfigure}[b]{0.96\textwidth}
         \centering
         \includegraphics[width=\textwidth]{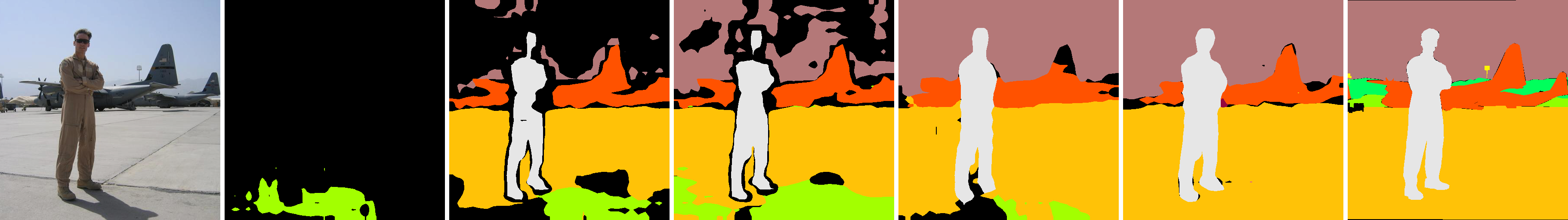}
     \end{subfigure}
        \vspace{-5pt}
     \caption{Qualitative results on the \textit{100-50} setting of the ADE20K dataset using different incremental methods. The image demonstrates the superiority of our approach on both new (\eg \textit{building}, \textit{floor}, \textit{table}) and old (\eg \textit{car}, \textit{wall}, \textit{person}) classes. From left to right: image, FT, LwF \cite{li2017learning}, ILT \cite{michieli2019incremental}, LwF-MC \cite{rebuffi2017icarl}, our method, and the ground-truth. Best viewed in color.}
        \label{fig:qualitative}
    \vspace{-10pt}
\end{figure*}

\subsection{ADE20K} \label{sec:ade}
ADE20K \cite{zhou2017scene} is a large-scale dataset that contains 150 classes. Differently from Pascal-VOC 2012, this dataset contains both stuff (\eg sky, building, wall)
and object classes. % but does not provide a class for the background.
We create the incremental datasets $\set T^t$ by splitting the whole dataset into disjoint image sets, without any constraint except ensuring a minimum number of images (\ie 50) where classes on $\set C^t$ have labeled pixels. % of the novel classes. 
Obviously, each $\set T^t$ provides annotations only for classes in $\set C^t$ while other classes (old or future) appear as background in the ground truth. %, that we introduce to implement the incremental protocol.
In Table \ref{tab:ade} we report the mean IoU obtained averaging the results on two different class orders: the order proposed by \cite{zhou2017scene} and a random one. In this experiments, we compare our approach with data-focused methods only (\ie LwF, LwF-MC, and ILT) due to their gap in performance with prior-focused ones.

%describe here the experiments and the results for ADE20K

\myparagraph{Single-step addition of 50 classes (\textit{100-50}).} In the first experiment, we initially train the network on 100 classes and we add the remaining 50 all at once.
From Table \ref{tab:ade} we can observe that FT is clearly a bad strategy on large scale settings since it completely forgets old knowledge. % classes. 
Using a distillation strategy enables the network to reduce the catastrophic forgetting: LwF obtains $21.1\%$ on past classes, ILT $22.9\%$, and LwF-MC $34.2\%$. Regarding new classes, LwF is the best strategy, exceeding LwF-MC by $18.9\%$ and ILT by $6.6\%$.
However, our method is far superior to all others, improving on the first classes and on the new ones. Moreover, we can observe that we are close to the joint training upper bound, especially considering new classes, where the gap with respect to it is only $0.3\%$.
In Figure \ref{fig:qualitative} we report some qualitative results which demonstrate the superiority of our method compared to the baselines.

\myparagraph{Multi-step addition of 50 classes (\textit{100-10}).} We then evaluate the performance on multiple incremental steps: we start from 100 classes and we add the remaining classes 10 by 10, resulting in 5 incremental steps. In Table \ref{tab:ade} we report the results on all sets of classes after the last learning step.
In this setting the performance of FT, LwF and ILT are very poor because they strongly suffers catastrophic forgetting. % Moreover, ILT exhibits a performance drop also on new classes, where it obtains $10.7\%$, $5.9\%$ worse than FT and LwF.
LwF-MC demonstrates a better ability to preserve knowledge on old classes, at the cost of a performance drop on new classes.
Again, our method achieves the best trade-off between learning new classes and preserving past knowledge, outperforming LwF-MC by $11.6\%$ considering all classes. %It is able to learn new classes obtaining results even superior to joint training (the incremental training is biased toward new classes), while it exhibit only a small drop in performance on previous steps.

\myparagraph{Three steps of 50 classes (\textit{50-50}).} Finally, in Table \ref{tab:ade} we analyze the performance on three sequential steps of 50 classes.
Previous \icl\ methods achieve different trade-offs between learning new classes and not forgetting old ones. LwF and ILT obtain a good score on new classes, but they forget old knowledge. On the contrary, LwF-MC preserves knowledge on the first 50 classes without being able to learn new ones.
Our method outperforms all the baselines by a large margin with a gap of $11.9\%$ on the best performing baseline, achieving the highest mIoU on every step. Remarkably, the highest gap 
%with the baselines 
is on the intermediate step, where there are classes that we must %there present 
 both learn incrementally and preserve from forgetting on the subsequent learning step. %where we must not only learn novel classes but also consolidate old knowledge.

\vspace{-4pt}
\section{Conclusions}
\vspace{-2pt}
{
% WHAT
%In this paper, 
We studied the incremental class learning problem for semantic segmentation, %Differently from previous works, we 
analyzing the realistic scenario where the new training set does not provide annotations for old classes, leading to the semantic shift of the background class and exacerbating the catastrophic forgetting problem.
% HOW
We address this issue by 
%revisiting standard incremental learning approaches, 
proposing a novel objective function and a classifiers' initialization strategy which allows our network to explicitly model the semantic shift of the background, effectively learning new classes without deteriorating its ability to recognize old ones.
%Our experimental 
Results show that
%effectiveness of 
our approach outperforms regularization-based \icl\ methods by a large margin, considering both small and large scale datasets. We hope that our problem formulation, our approach and our extensive comparison with previous methods will encourage future works on this novel research topic.
%In our experimental analysis, we benchmark the performances of parameter-based regularization and distillation-based \icl\ methods using both small and large scale datasets. Our experimental results show the effectiveness of our approach which outperforms all the baselines by a large margin. We hope that our problem formulation, our approach and our extensive comparison with previous methods will encourage future works on this novel research topic.

}
{\small
\bibliographystyle{ieee_fullname}
\bibliography{egbib}
}

\end{document}

%%%%%%%%% TITLE
\title{Modeling the Background for Incremental Learning in Semantic Segmentation\\[5pt] \LARGE Supplementary Material}

\author{
Fabio Cermelli$^{1,2}$, Massimiliano Mancini$^{2,3,4}$, Samuel Rota Bul\`o$^{5}$, Elisa Ricci$^{3,6}$, Barbara Caputo$^{1,2}$\\
$^1$Politecnico di Torino, $^2$Italian Institute of Technology, $^3$Fondazione Bruno Kessler,\\ $^4$Sapienza University of Rome,
$^5$Mapillary Research, $^6$University of Trento \\
{\tt\small \{fabio.cermelli, barbara.caputo\}@polito.it, mancini@diag.uniroma1.it,} \\
{\tt\small samuel@mapillary.com, eliricci@fbk.eu}
}

\maketitle

\section{Qualitative results}
We provide qualitative results of our method and the baselines on %\textit{randomly sampled} No dai, non possiamo permettercelo...
images of the validation sets of ADE20k \cite{zhou2017scene} and Pascal-VOC 2012 \cite{pascal-voc-2012} for different incremental learning scenarios. The results are shown in the accompanying video (\ie \textit{video.mp4}), together with an illustration of the background shift problem. As the video shows, the naive strategy of tuning network parameters (FT) for each step independently fails, forgetting all the old classes. Similarly, since the background shift exacerbates catastrophic forgetting, previous incremental learning methods that do not model the background (LwF \cite{li2017learning}, LwF-MC \cite{rebuffi2017icarl}, ILT \cite{michieli2019incremental}) are not able to either maintain previous knowledge or learn novel concepts. Our method (MiB) instead can both learn new classes while not forgetting old ones, showing its effectiveness in modeling the background shift for incremental learning in semantic segmentation.

\section{How should we use the background?}
As highlighted in the main paper,  an important design choice for incremental learning in semantic segmentation is how to use the background. In particular, since the background class is present both in old and new classes, it can be considered either in the supervised cross-entropy loss, in the distillation component or in both. For our method and all the baselines (LwF \cite{li2017learning}, Lwf-MC \cite{rebuffi2017icarl}, ILT \cite{michieli2019incremental}), we considered the latter case (\ie background in both). However, a natural question arises on how different choices for the background would impact the final results. In this section we investigate this point. 

We start from the LwF-MC \cite{rebuffi2017icarl} baseline, since it is composed of multiple binary classifiers and allows to easy decouple modifications on the background from the other classes. We then test two variants: 
 \begin{itemize}
   % \item \textbf{LwF-MC}, which is the one reported in the paper, uses a weighted combination among two binary cross-entropy losses, using the background on both the ground truth and the probabilities given by $f_{\theta^{t-1}}$.
    \item \textbf{LwF-MC-D} ignores the background in the classification loss, using as %BCE 
    target for the background the probability given by $f_{\theta^{t-1}}$.
    \item \textbf{LwF-MC-C} ignores the background in the distillation loss, using only the supervised signal from the ground-truth.
\end{itemize}
In Table \ref{tab:lwf-voc} and \ref{tab:lwf-ade} we report the results of the two variants for the overlapped scenarios of the Pascal VOC dataset and the \textit{50-50} scenario of ADE20K respectively. Together with the two variants, we report the results of our method (MiB), the offline training upper-bound (Joint) and the LwF-MC version employed in the paper which uses the background in both binary cross-entropy and distillation, blending the two components with a hyper-parameter. 

As the tables show, the three variants of Lwf-MC exhibit different trade-offs among learning new knowledge and remembering the past one.
In particular, LwF-MC-C learns very well new classes, being always the most performing variant on the last incremental step. However, it suffers a significant drop in the old knowledge, showing its inability to tackle the catastrophic forgetting problem.

LwF-MC-D shows the opposite trend. It maintains very well the old knowledge, being the best variant in old classes for every setting. However, it is very intransigent \cite{chaudhry2018riemannian} \ie it is not able to correctly learn new classes, thus obtaining the worst performances on them. 

As expected, LwF-MC which considers the background in both cross-entropy and distillation achieves a trade-off among learning new knowledge, as in LwF-MC-C, while preserving the old one, as in LwF-MC-D. %We highlight also that the fact that, performance wise, LwF-MC-D surpasses LwF-MC is linked to two aspects (i) the number of old classes is larger in the VOC scenarios, thus preserving the old knowledge leads to higher results even if nothing is learned for the new ones and (ii) the cross-validation protocol of \cite{} works in favor of hyper-parameters which do not damage performances on new learned classes.

As the tables show, our MiB approach models the background more effectively, achieving the best trade-off among learning new knowledge and preserving old concepts. In particular, our method is the best by a margin in all scenarios for the new classes, while for old ones it is either better or comparable to the performance of the intransigent LwF-MC-D method. The only scenarios where it shows lower performances are the multi-step ones. Indeed in these scenarios, the multiple learning episodes make preserving old knowledge harder, and an intransigent method is less prone to forgetting since it is biased to old classes. However, the intransigence is not the right solution if the number of old and new classes are balanced, as in the \textit{50-50} scenario of ADE20k, since the overall performances will be damaged.

% Finally,we highlight that the cross-validation protocol of \cite{de2019continual} works in favor of not damaging the ability to learn new classes, thus using a different protocol might raise the performances of LwF-MC on old classes.

%Despite LwF-MC-D obtains better overall results, we consider LwF-MC the most performing implementation since it exhibits the best trade-off between learning and forgetting, obtaining performance in between previous methods. In particular, it is better than LwF-MC-C on old classes ($64.4\%$ on \textit{19-1}, $58.1\%$ on \textit{15-5}, and $6.4\%$ on \textit{15-1}) and than LwF-MC-D on new ones ($13.3\%$ on \textit{19-1}, $35.0\%$ on \textit{15-5}, and $8.4\%$ on \textit{15-1}).

%In this section, we compare three different versions of the distillation employed by LwF-MC \cite{rebuffi2017icarl}, considering the different use of the background class.
%LwF-MC is based on multiple binary classifiers, with the target labels defined using the ground truth for novel classes (\ie $\set C^t$) and the probabilities given by the old model for the old ones (\ie $\set Y^{t-1}$). Since the background class is both in $\set C^t$ and $\set Y^{t-1}$, we can choose where to use it. 
%In particular, we report three different implementations:

\begin{table}[t]
\small
\centering
\setlength{\tabcolsep}{2pt} % Default value: 6pt
\caption{Comparison of different implementations of LwF-MC on the Pascal-VOC 2012 \textit{overlapped} setup.}
\label{tab:lwf-voc}
\begin{tabular}{l||rr|r||rr|r||rr|r}
\multicolumn{1}{c}{}& \multicolumn{3}{c}{\textbf{19-1}}          & \multicolumn{3}{c}{\textbf{15-5}}          & \multicolumn{3}{c}{\textbf{15-1}} \\
\textbf{Method} & \it{1-19} & \it{20} & \it{all} & \it{1-15} & \it{16-20} & \it{all} & \it{1-15}  & \it{16-20}     & \it{all}     \\ \hline
%FT         & 6.8  & 12.9 & 7.1  & 2.1  & 33.1 & 9.8  & 0.2  & 1.8  & 0.6  \\ \hline
LwF-MC-C & 44.6 & \underline{17.6} & 43.2 & 41.6 & \underline{42.2} & 41.8 & 4.4  & \underline{8.6}  & 5.4  \\
LwF-MC   & 64.4 & 13.3 & 61.9 & 58.1 & 35.0 & 52.3 & 6.4  & 8.4  & 6.9  \\
LwF-MC-D & \textbf{71.3} & 3.6  & \textbf{68.0} & \underline{73.7} & 21.0 & \underline{60.5} & \textbf{41.1} & 3.1  & \textbf{31.6} \\ 
MiB   & \underline{70.2}      & \textbf{22.1}      & \underline{67.8}      & \textbf{75.5}      & \textbf{49.4}      & \textbf{69.0}      & \underline{35.1}       & \textbf{13.5} & \underline{29.7}\\
\hline
Joint      & 77.4 & 78.0 & 77.4 & 79.1 & 72.6 & 77.4 & 79.1 & 72.6 & 77.4 

\end{tabular}
\end{table}

% \myparagraph{Pascal-VOC 2012}
% In Table \ref{tab:lwf-voc}, we compare the different implementations on the overlapped setup of the Pascal-VOC 2012 dataset, considering the addition of one class (\textit{19-1}), five classes at once (\textit{15-5}), and five classes sequentially (\textit{15-1}).
% The three methods exhibit different trade-offs among learning the new and remembering the past.
% LwF-MC-C learns very well new classes, being always the most performing method on them ($17.6\%$ on \textit{19-1}, $42.2\%$ on \textit{15-5}, and $8.6\%$ on \textit{15-1}). However, it suffers a significant drop on the initial classes, which limits the overall performance.
% LwF-MC-D shows the opposite trade-off. It remembers very well past knowledge, being the best on them in every setting ($71.3\%$ on \textit{19-1}, $73.7\%$ on \textit{15-5}, and $41.4\%$ on \textit{15-1}), but it is not able to learn correctly new classes, demonstrating the worst performance on them.
% Despite LwF-MC-D obtains better overall results, we consider LwF-MC the most performing implementation since it exhibits the best trade-off between learning and forgetting, obtaining performance in between previous methods. In particular, it is better than LwF-MC-C on old classes ($64.4\%$ on \textit{19-1}, $58.1\%$ on \textit{15-5}, and $6.4\%$ on \textit{15-1}) and than LwF-MC-D on new ones ($13.3\%$ on \textit{19-1}, $35.0\%$ on \textit{15-5}, and $8.4\%$ on \textit{15-1}).

\begin{table}[]
\small
\centering
\caption{Comparison of different implementations of LwF-MC on the \textit{50-50} setting of the ADE20K dataset}
\label{tab:lwf-ade}
\begin{tabular}{l||rrr|r}
\textbf{Method}& \textit{1-50} & \textit{51-100} & \textit{101-150} & \textit{all}  \\ \hline
%FT        & 0.0  & 0.0    & 22.0    & 7.3  \\ \hline
LwF-MC-C  & 8.0  & 7.2    & \underline{19.3}    & 11.5 \\
LwF-MC    & 27.8 & 7.0    & 10.4    & 15.1 \\
LwF-MC-D  & \textbf{39.1} & \underline{10.9}   & 6.7     & \underline{18.7} \\ 
MiB & \underline{35.5} & \textbf{22.2}   & \textbf{23.6}    & \textbf{27.0} \\\hline
Joint     & 51.1 & 38.3   & 28.2    & 38.9
\end{tabular}
\end{table}

% \myparagraph{ADE20K}
% After evaluating the methods on Pascal-VOC 2012, we were interested in comparing them also on a large-scale scenario. We then analyzed the behavior of the three implementation on the ADE20K dataset, considering an initial training of 50 classes, followed by two learning step of 50 classes. 
% The results, showed in Table \ref{tab:lwf-ade} indicates a behavior consistent with Pascal-VOC 2012. In particular, LwF-MC-C suffers a significant forgetting of previous knowledge, achieving $8.0\%$ on the first 50 classes (\ie $\set C^1$) and $7.2\%$ on the classes from the second step (\ie $\set C^2$).
% LwF-MC-D demonstrate a good ability to maintain knowledge, but it struggles to learn new concept over time, achieving $6.7\%$ on the last 50 classes (\ie $\set C^3$).
% Finally, LwF-MC demonstrates the best trade-off between learning and forgetting, achieving $8.0\%$ on $\set C^1$,  $7.0\%$ on $\set C^2$, and $10.4\%$ on $\set C^3$.

%\section{Per class results on Pascal-VOC 2012}
%To provide the reader a more detailed analysis of the results, in this section we report the result in term of mean IoU per each class.

\section{Per class results on Pascal-VOC 2012}
\begin{table*}[t]
\centering
\caption{Per Class Mean IoU on 19-1 setting of Pascal-VOC 2012. disjoint setup}
\resizebox{\textwidth}{!}{
\label{tab:voc-19-1-d}
\begin{tabular}{l|rrrrrrrrrrrrrrrrrrr|r||r|r}
\textbf{Method} & aero & bike & bird & boat & bottle & bus  & car  & cat  & chair & cow  & table & dog  & horse & mbike & persn & plant & sheep & sofa & train & tv   & \textbf{1-19} & \textbf{all} \\ \hline
FT              & 11.9 & 2.1  & 1.1  & 11.6 & 4.8    & 6.9  & 13.5 & 0.2  & 0.0   & 3.8  & 14.4  & 0.5  & 1.5   & 4.7   & 0.0   & 15.8  & 2.8   & 1.8  & 13.5  & 12.3 & 5.8  & 6.2          \\
PI~\cite{zenke2017continual}              & 22.3 & 1.9  & 3.4  & 4.9  & 2.1    & 10.6 & 8.5  & 0.1  & 0.1   & 3.1  & 12.8  & 0.2  & 3.8   & 4.6   & 0.0   & 10.0  & 5.0   & 1.1  & 8.5   & 14.1 & 5.4  & 5.9          \\
EWC~\cite{kirkpatrick2017overcoming}             & 50.7 & 7.7  & 21.0 & 24.1 & 21.8   & 35.8 & 43.9 & 11.6 & 2.0   & 27.0 & 21.1  & 23.0 & 18.7  & 19.4  & 1.5   & 27.8  & 41.5  & 5.6  & 37.4  & 16.0 & 23.2 & 22.9         \\
RW~\cite{chaudhry2018riemannian}             & 45.8 & 5.3  & 15.1 & 22.8 & 17.8   & 28.9 & 40.9 & 7.5  & 1.3   & 22.4 & 20.3  & 14.5 & 13.7  & 16.3  & 0.8   & 25.3  & 31.8  & 4.8  & 33.3  & 15.7 & 19.4 & 19.2         \\
LwF~\cite{li2017learning}            & 28.1 & 40.5 & 53.1 & 38.8 & 47.4   & 46.4 & 63.6 & 83.5 & 35.8  & 60.1 & 48.8  & 76.5 & 65.3  & 67.1  & 83.2  & 50.2  & 61.2  & 42.5 & 14.2  & 9.1  & 53.0 & 50.8         \\
LwF-MC~\cite{rebuffi2017icarl}         & {79.4} & \textbf{41.3} & 75.6 & 47.9 & 51.0   & 69.6 & 75.4 & 78.5 & 35.1  & 66.6 & 49.0  & 72.7 & 73.8  & 71.6  & \textbf{84.9}  & \textbf{57.5}  & 67.7  & 42.7 & 56.8  & 13.2 & 63.0 & 60.5         \\
ILT~\cite{michieli2019incremental}            & \textbf{83.7} & 40.8 & 80.8 & \textbf{59.1} & 58.4   & 77.6 & \textbf{82.4} & 82.3 & 38.9  & 81.7 & 50.8  & 84.8 & \textbf{86.6}  & \textbf{81.0}  & 83.3  & 56.4  & 82.2  & 43.8 & 57.5  & 16.4 & 69.1 & 66.4         \\
MiB             & 78.0 & 40.5 & \textbf{85.7} & 51.6 & \textbf{64.4}   & \textbf{79.1} & {77.8} & \textbf{89.9} & \textbf{39.2} & \textbf{82.3} & \textbf{55.4}  & \textbf{86.2} & 82.7  & 72.2  & 83.6  & 56.6  & \textbf{86.2} & \textbf{45.1} & \textbf{65.0}  & \textbf{25.6} & \textbf{69.6} & \textbf{67.4}         \\ \hline
Joint           & 90.2 & 42.2 & 89.5 & 69.1 & 82.3   & 92.5 & 90.0 & 94.2 & 39.2  & 87.6 & 56.4  & 91.2 & 86.8  & 88.0  & 86.8  & 62.3  & 88.4  & 49.5 & 85.0  & 78.0 & 77.4 & 77.4        
\end{tabular}}
\end{table*}

\begin{table*}[t]
\centering
\caption{Per Class Mean IoU on 19-1 setting of Pascal-VOC 2012. overlapped setup}
\resizebox{\textwidth}{!}{
\label{tab:voc-19-1-o}
\begin{tabular}{l|rrrrrrrrrrrrrrrrrrr|r||r|r}
\textbf{Method} & aero & bike & bird & boat & bottle & bus  & car  & cat  & chair & cow  & table & dog  & horse & mbike & persn & plant & sheep & sofa & train & tv   & \textbf{1-19} & \textbf{all} \\ \hline
FT     & 23.7 & 1.9  & 1.5  & 9.3  & 6.9    & 16.9 & 8.5  & 0.0  & 0.0   & 9.5  & 5.3      & 0.1  & 2.9   & 8.8   & 0.0    & 15.1  & 1.0   & 0.7  & 16.0  & 12.9 & 6.8  & 7.1  \\
PI~\cite{zenke2017continual}     & 33.1 & 4.1  & 3.6  & 10.5 & 8.4    & 14.7 & 13.3 & 0.0  & 0.1   & 2.4  & 4.7      & 0.1  & 3.3   & 7.9   & 0.0    & 14.7  & 0.8   & 2.7  & 17.8  & 14.0 & 7.5  & 7.8  \\
EWC~\cite{kirkpatrick2017overcoming}   & 60.7 & 14.8 & 21.2 & 33.8 & 36.9   & 54.4 & 45.6 & 2.6  & 1.4   & 33.0 & 13.3     & 19.1 & 23.8  & 39.2  & 2.2    & 34.6  & 21.8  & 6.4  & 47.1  & 14.0 & 26.9 & 26.3 \\
RW~\cite{chaudhry2018riemannian}     & 57.5 & 12.1 & 15.4 & 29.6 & 32.9   & 50.7 & 40.0 & 1.3  & 0.8   & 30.7 & 10.7     & 12.6 & 18.6  & 32.9  & 0.8    & 30.7  & 17.5  & 5.5  & 42.7  & 14.2 & 23.3 & 22.9 \\
LwF~\cite{li2017learning}    & 36.6 & 35.1 & 62.0 & 32.9 & 47.5   & 31.6 & 51.5 & 77.9 & 36.5  & 67.7 & 44.3     & 71.4 & 68.6  & 66.2  & 82.2   & 49.6  & 58.7  & 41.1 & 11.9  & 8.5  & 51.2 & 49.1 \\
LwF-MC~\cite{rebuffi2017icarl} & 67.2 & 37.9 & 77.8 & 40.6 & 57.0   & 54.5 & 77.4 & 88.4 & 37.2  & 76.8 & 49.1     & 83.4 & 82.3  & 71.0  & \textbf{85.2}   & 55.6  & 81.9  & \textbf{46.0} & 54.9  & 13.3 & 64.4 & 61.9 \\
ILT~\cite{michieli2019incremental}    & \textbf{87.2} & \textbf{39.0} & 80.6 & \textbf{53.5} & 57.0   & 80.3 & 76.0 & 74.3 & 37.6  & 81.1 & 44.6     & 83.1 & \textbf{84.4}  & 81.6  & 82.4   & 54.5  & 82.7  & 38.9 & 56.1  & 12.3 & 67.1 & 64.4 \\
MiB    & 78.1 & 36.2 & \textbf{86.8} & 49.4 & \textbf{72.7}   & \textbf{80.8} & \textbf{78.2} & \textbf{90.8} & \textbf{38.3}  & \textbf{82.0} & \textbf{51.9}     & \textbf{86.7} & 82.8  & \textbf{76.9}  & 83.8   & \textbf{58.8}  & \textbf{84.4}  & 45.7 & \textbf{68.5}  & \textbf{22.1} & \textbf{70.2} & \textbf{67.8} \\ \hline
Joint  & 90.2 & 42.2 & 89.5 & 69.1 & 82.3   & 92.5 & 90.0 & 94.2 & 39.2  & 87.6 & 56.4     & 91.2 & 86.8  & 88.0  & 86.8   & 62.3  & 88.4  & 49.5 & 85.0  & 78.0 & 77.4 & 77.4
\end{tabular}}
\end{table*}

\begin{table*}[t]
\centering
\caption{Per Class Mean IoU on 15-5 setting of Pascal-VOC 2012. disjoint setup}
\resizebox{\textwidth}{!}{
\label{tab:voc-15-5-d}
\begin{tabular}{l|rrrrrrrrrrrrrrr|rrrrr||r|r|r}
\textbf{Method} & aero & bike & bird & boat & bottle & bus  & car  & cat  & chair & cow  & table & dog  & horse & mbike & persn & plant & sheep & sofa & train & tv   & \textbf{1-15} & \textbf{16-20} & \textbf{all} \\ \hline
FT     & 6.1  & 0.0  & 0.2  & 8.3  & 0.1    & 0.0  & 0.1  & 0.0  & 0.0   & 0.0  & 0.0   & 0.0  & 1.8   & 0.0   & 0.0   & 24.6  & 24.3  & 36.2 & 32.5  & 50.2 & 1.1  & 33.6  & 9.2  \\
PI~\cite{zenke2017continual}     & 8.8  & 0.0  & 0.2  & 10.5 & 0.0    & 0.0  & 0.1  & 0.0  & 0.0   & 0.0  & 0.0   & 0.0  & 0.4   & 0.0   & 0.0   & 25.6  & 24.7  & 34.3 & 34.1  & 52.0 & 1.3  & 34.1  & 9.5  \\
EWC~\cite{kirkpatrick2017overcoming}   & 58.8 & 4.1  & 56.4 & 46.2 & 44.4   & 4.3  & 67.4 & 3.6  & 2.3   & 14.8 & 10.3  & 12.4 & 51.6  & 20.4  & 2.9   & 28.8  & 32.2  & 35.6 & 35.5  & 56.3 & 26.7 & 37.7  & 29.4 \\
RW~\cite{chaudhry2018riemannian}    & 51.1 & 1.5  & 36.9 & 42.9 & 27.5   & 2.1  & 47.4 & 1.1  & 1.2   & 6.1  & 5.3   & 3.1  & 31.2  & 10.5  & 1.0   & 27.7  & 29.8  & 35.7 & 34.7  & \textbf{56.6} & 17.9 & 36.9  & 22.7 \\
LwF~\cite{li2017learning}   & 63.1 & 40.1 & 72.4 & 52.1 & 67.0   & 6.7  & 80.3 & 84.2 & 31.1  & 5.7  & 51.3  & 82.0 & 75.0  & 79.4  & 85.6  & 35.3  & 27.1  & 37.0 & 37.0  & 50.5 & 58.4 & 37.4  & 53.1 \\
LwF-MC~\cite{rebuffi2017icarl}& 78.1 & \textbf{42.3} & 78.9 & 62.1 & 78.6   & 47.3 & 84.6 & 89.1 & 35.0  & 26.2 & 50.5  & 86.6 & 77.6  & \textbf{84.9}  & \textbf{86.0}  & 35.0  & 35.2  & \textbf{40.8} & 49.2  & 45.9 & 67.2 & 41.2  & 60.7 \\
ILT~\cite{michieli2019incremental}   & 79.4 & 42.0 & 80.5 & 63.9 & \textbf{80.4}   & 12.8 & \textbf{86.0} & 90.2 & 30.7  & 6.7  & \textbf{53.3}  & 83.2 & 73.0  & 80.7  & 85.0  & \textbf{36.9}  & 29.9  & 36.8 & 38.3  & 55.7 & 63.2 & 39.5  & 57.3 \\
MiB    & \textbf{84.4} & 39.4 & \textbf{87.5} & \textbf{65.2} & 77.8   & \textbf{61.0} & \textbf{86.0} & \textbf{90.9 }& \textbf{35.3}  & \textbf{60.3} & 53.0  & \textbf{88.2} & \textbf{80.4}  & 82.4  & 85.3  & 28.7  & \textbf{46.0}  & 34.7 & \textbf{54.4}  & 52.7 & \textbf{71.8} & \textbf{43.3}  & \textbf{64.7} \\ \hline
Joint  & 90.2 & 42.2 & 89.5 & 69.1 & 82.3   & 92.5 & 90.0 & 94.2 & 39.2  & 87.6 & 56.4  & 91.2 & 86.8  & 88.0  & 86.8  & 62.3  & 88.4  & 49.5 & 85.0  & 78.0 & 79.1 & 72.6  & 77.4
\end{tabular}}
\end{table*}

\begin{table*}[t]
\centering
\caption{Per Class Mean IoU on 15-5 setting of Pascal-VOC 2012. overlapped setup}
\resizebox{\textwidth}{!}{
\label{tabvoc-:15-5-o}
\begin{tabular}{l|rrrrrrrrrrrrrrr|rrrrr||r|r|r}
\textbf{Method} & aero & bike & bird & boat & bottle & bus  & car  & cat  & chair & cow  & table & dog  & horse & mbike & persn & plant & sheep & sofa & train & tv   & \textbf{1-15} & \textbf{16-20} & \textbf{all} \\ \hline
FT     & 13.4 & 0.1  & 0.0  & 15.6 & 0.8    & 0.0  & 0.3  & 0.0  & 0.0   & 0.0  & 0.0   & 0.0  & 0.9   & 0.0   & 0.0   & 30.9  & 21.6  & 32.8 & 34.9  & 45.1 & 2.1  & 33.1  & 9.8  \\
PI~\cite{zenke2017continual}     & 7.8  & 0.0  & 0.0  & 12.9 & 0.3    & 0.0  & 0.3  & 0.0  & 0.0   & 0.0  & 0.0   & 0.0  & 2.7   & 0.0   & 0.0   & 33.2  & 22.2  & 33.2 & 36.1  & 42.0 & 1.6  & 33.3  & 9.5  \\
EWC~\cite{kirkpatrick2017overcoming}   & 67.3 & 12.8 & 50.5 & 52.9 & 35.0   & 24.7 & 41.7 & 1.2  & 1.0   & 9.8  & 5.7   & 3.7  & 42.9  & 15.4  & 0.6   & 31.8  & 26.3  & 32.1 & 42.0  & 45.0 & 24.3 & 35.5  & 27.1 \\
RW~\cite{chaudhry2018riemannian}    & 61.2 & 6.7  & 33.8 & 48.1 & 24.4   & 9.3  & 22.3 & 0.3  & 0.5   & 3.5  & 0.2   & 1.1  & 31.8  & 6.4   & 0.1   & 32.1  & 25.8  & 31.9 & 38.7  & 45.9 & 16.6 & 34.9  & 21.2 \\
LwF~\cite{li2017learning}   & 64.5 & 40.2 & 72.8 & 56.9 & 57.3   & 9.5  & 82.6 & 88.6 & 33.2  & 8.9  & 48.4  & 81.9 & 75.0  & 78.2  & 84.9  & 34.7  & 27.8  & 33.1 & 39.6  & 48.0 & 58.9 & 36.6  & 53.3 \\
LwF-MC~\cite{rebuffi2017icarl}& 60.6 & 38.9 & 74.7 & 41.6 & 67.2   & 10.8 & 81.4 & 88.8 & \textbf{38.7}  & 4.3  & 47.4  & 82.2 & 69.9  & 78.9  & \textbf{85.8}  & 28.4  & 28.5  & \textbf{34.1} & 36.4  & 47.8 & 58.1 & 35.0  & 52.3 \\
ILT~\cite{michieli2019incremental}   & 77.4 & \textbf{40.3} & 78.9 & 61.9 & 78.7   & 53.5 & 86.1 & 88.7 & 33.8  & 15.9 & 51.1  & 83.2 & 80.2  & 79.8  & 85.0  & \textbf{39.5}  & 30.9  & 31.0 & 49.3  & 52.6 & 66.3 & 40.6  & 59.9 \\
MiB    & \textbf{86.6} & 39.3 & \textbf{88.9} & \textbf{66.1} & \textbf{80.8}   & \textbf{86.6 }& \textbf{90.1} & \textbf{92.5} & 38.0  & \textbf{64.6} & \textbf{56.4}  & \textbf{89.6} & \textbf{80.5}  & \textbf{86.5}  & 85.7  & 30.2  & \textbf{52.9}  & 31.3 & \textbf{73.2}  & \textbf{59.5} & \textbf{75.5} & \textbf{49.4}  & \textbf{69.0} \\ \hline
Joint  & 90.2 & 42.2 & 89.5 & 69.1 & 82.3   & 92.5 & 90.0 & 94.2 & 39.2  & 87.6 & 56.4  & 91.2 & 86.8  & 88.0  & 86.8  & 62.3  & 88.4  & 49.5 & 85.0  & 78.0 & 79.1 & 72.6  & 77.4
\end{tabular}}
\end{table*}

\begin{table*}[t]
\centering
\caption{Per Class Mean IoU on 15-1 setting of Pascal-VOC 2012. disjoint setup}
\resizebox{\textwidth}{!}{
\label{tab:voc-15-1-d}
\begin{tabular}{l|rrrrrrrrrrrrrrr|r|r|r|r|r||r|r|r}
\textbf{Method} & aero & bike & bird & boat & bottle & bus  & car  & cat  & chair & cow  & table & dog  & horse & mbike & persn & plant & sheep & sofa & train & tv   & \textbf{1-15} & \textbf{16-20} & \textbf{all} \\ \hline
FT     & 0.3  & 0.0  & 0.0  & 2.5  & 0.0    & 0.0  & 0.0  & 0.0  & 0.0   & 0.0  & 0.0   & 0.0  & 0.0   & 0.0   & 0.0   & 0.0   & 0.0   & 0.0  & 0.0   & 8.8  & 0.2  & 1.8   & 0.6  \\
PI~\cite{zenke2017continual}     & 0.0  & 0.0  & 0.0  & 0.0  & 0.0    & 0.0  & 0.0  & 0.0  & 0.0   & 0.0  & 0.0   & 0.0  & 0.0   & 0.0   & 0.0   & 0.0   & 0.0   & 0.0  & 0.3   & 8.6  & 0.0  & 1.8   & 0.4  \\
EWC~\cite{kirkpatrick2017overcoming}   & 0.0  & 0.0  & 0.0  & 1.0  & 0.0    & 0.0  & 0.0  & 0.0  & 0.0   & 0.0  & 3.6   & 0.0  & 0.0   & 0.0   & 0.0   & 0.0   & 0.0   & 7.3  & 7.0   & 7.4  & 0.3  & 4.3   & 1.3  \\
RW~\cite{chaudhry2018riemannian}    & 0.0  & 0.0  & 0.0  & 0.2  & 0.0    & 0.0  & 0.0  & 0.0  & 0.0   & 0.0  & 2.2   & 0.0  & 0.0   & 0.0   & 0.0   & 0.0   & 0.0   & 8.1  & 10.5  & 8.2  & 0.2  & 5.4   & 1.5  \\
LwF~\cite{li2017learning}   & 0.0  & 0.0  & 0.0  & 0.0  & 0.6    & 0.0  & 0.0  & 0.0  & 0.0   & 0.0  & 0.0   & 0.0  & 0.0   & 0.0   & 10.7  & 0.0   & 0.0   & 1.9  & 8.2   & 7.9  & 0.8  & 3.6   & 1.5  \\
LwF-MC~\cite{rebuffi2017icarl}& 0.0  & 6.3  & 0.8  & 0.0  & 1.1    & 0.0  & 0.1  & 0.3  & 0.0   & 0.0  & 0.0   & 0.0  & 0.2   & 0.0   & 59.0  & 0.0   & 9.5   & 2.9  & 11.9  & \textbf{11.0} & 4.5  & 7.0   & 5.2  \\
ILT~\cite{michieli2019incremental}   & 3.7  & 0.0  & 2.9  & 0.0  & 12.8   & 0.0  & 0.0  & 0.1  & 0.0   & 0.0  & \textbf{21.2}  & 0.1  & 0.4   & 0.6   & 13.6  & 0.0   & 0.0   & 11.6 & 8.3   & 8.5  & 3.7  & 5.7   & 4.2  \\
MiB    & \textbf{53.6} & \textbf{38.9} & \textbf{53.6} & \textbf{17.7} & \textbf{62.7}   & \textbf{36.5} & \textbf{71.2} & \textbf{60.1} & \textbf{1.1}   & \textbf{35.2} & 8.1   & \textbf{57.6} & \textbf{55.0}  & \textbf{62.1}  & \textbf{79.4}  & \textbf{10.2}  & \textbf{14.2}  & \textbf{11.9} & \textbf{18.2} & 10.1 & \textbf{46.2} & \textbf{12.9}  & \textbf{37.9} \\ \hline
Joint  & 90.2 & 42.2 & 89.5 & 69.1 & 82.3   & 92.5 & 90.0 & 94.2 & 39.2  & 87.6 & 56.4  & 91.2 & 86.8  & 88.0  & 86.8  & 62.3  & 88.4  & 49.5 & 85.0  & 78.0 & 79.1 & 72.6  & 77.4
\end{tabular}}
\end{table*}

\begin{table*}[t]
\centering
\caption{Per Class Mean IoU on 15-1 setting of Pascal-VOC 2012. overlapped setup}
\resizebox{\textwidth}{!}{
\label{tab:voc-15-1-o}
\begin{tabular}{l|rrrrrrrrrrrrrrr|r|r|r|r|r||r|r|r}
\textbf{Method} & aero & bike & bird & boat & bottle & bus  & car  & cat  & chair & cow  & table & dog  & horse & mbike & persn & plant & sheep & sofa & train & tv   & \textbf{1-15} & \textbf{16-20} & \textbf{all} \\ \hline
FT     & 2.6  & 0.0  & 0.0  & 0.7  & 0.0    & 0.1  & 0.0  & 0.0  & 0.0   & 0.0  & 0.0   & 0.0  & 0.0   & 0.0   & 0.0   & 0.0   & 0.0   & 0.0  & 0.0   & 9.2  & 0.2  & 1.8   & 0.6  \\
PI~\cite{zenke2017continual}     & 0.0  & 0.0  & 0.0  & 0.0  & 0.0    & 0.0  & 0.0  & 0.0  & 0.0   & 0.0  & 0.0   & 0.0  & 0.0   & 0.0   & 0.0   & 0.0   & 0.0   & 0.0  & 0.2   & 9.1  & 0.0  & 1.8   & 0.5  \\
EWC~\cite{kirkpatrick2017overcoming}   & 0.0  & 0.0  & 0.0  & 1.0  & 0.0    & 0.0  & 0.0  & 0.0  & 0.0   & 0.0  & 3.6   & 0.0  & 0.0   & 0.0   & 0.0   & 0.0   & 0.0   & 7.3  & 7.0   & 7.4  & 0.3  & 4.3   & 1.3  \\
RW~\cite{chaudhry2018riemannian}    & 0.1  & 0.0  & 0.0  & 0.0  & 0.0    & 0.0  & 0.0  & 0.0  & 0.0   & 0.0  & 0.0   & 0.0  & 0.0   & 0.0   & 0.0   & 0.0   & 0.0   & 8.7  & {11.2}  & 6.3  & 0.0  & 5.2   & 1.3  \\
LwF~\cite{li2017learning}   & 3.7  & 0.1  & 0.0  & 2.5  & 0.2    & 0.0  & 0.0  & 0.0  & 0.0   & 0.0  & 0.0   & 0.0  & 0.1   & 0.0   & 9.0   & 0.0   & 0.0   & 1.6  & 8.9   & 8.8  & 1.0  & 3.9   & 1.8  \\
LwF-MC~\cite{rebuffi2017icarl}& 0.0  & 7.2  & 5.2  & 0.0  & 25.5   & 0.0  & 0.0  & 0.0  & 0.0   & 0.0  & 0.0   & 0.0  & 1.2   & 1.3   & 56.2  & 0.0   & 4.9   & 0.2  & 8.6   & \textbf{28.2} & 6.4  & 8.4   & 6.9  \\
ILT~\cite{michieli2019incremental}   & 20.0 & 0.0  & 3.2  & 6.3  & 2.3    & 0.0  & 0.0  & 0.0  & \textbf{0.3}   & 5.1  & \textbf{19.0}  & 0.0  & 9.1   & 0.0   & 8.7   & 0.0   & 0.0   & \textbf{21.0} & 9.9   & 8.1  & 4.9  & 7.8   & 5.7  \\
MiB    & \textbf{31.3} &\textbf{ 25.4} & \textbf{26.7} & \textbf{26.9} & \textbf{46.1}   & \textbf{31.0} & \textbf{63.6} & \textbf{52.8} & 0.1   & \textbf{11.0} & 9.4   & \textbf{52.4} & \textbf{41.2}  & \textbf{28.1}  & \textbf{80.7}  & \textbf{17.6}  & \textbf{13.1}  & 15.3 & \textbf{15.3}  & 6.2  & \textbf{35.1} & \textbf{13.5}  & \textbf{29.7} \\ \hline
Joint  & 90.2 & 42.2 & 89.5 & 69.1 & 82.3   & 92.5 & 90.0 & 94.2 & 39.2  & 87.6 & 56.4  & 91.2 & 86.8  & 88.0  & 86.8  & 62.3  & 88.4  & 49.5 & 85.0  & 78.0 & 79.1 & 72.6  & 77.4
\end{tabular}}
\end{table*}

From Table \ref{tab:voc-19-1-d} to \ref{tab:voc-15-1-o}, we report the results for all classes of the Pascal-VOC 2012 dataset. As the tables show, MiB achieves the best results in the majority of classes (i.e. at least 14/20 in the 19-1 scenarios, 13/20 in the 15-5 and 16/20 in the 15-1 ones) being either the second best or comparable to the top two in all the others. Remarkable cases are the ones where we learn classes that are either similar in appearance (\eg bus and train) or appear in similar contexts (\eg sheep and cow): for those pairs, our model outperforms the competitors by a margin in both old classes (\ie bus and cow in the 15-5 and 15-1 scenarios) and new ones (\ie sheep and train). These results show the capability of MiB to not only learn new knowledge while preserving the old one, but also to learn discriminative features for difficult cases during different learning steps.

\section{Validation protocol and \hypers}
In this work, we follow the protocol of \cite{de2019continual} for setting the \hypers in continual learning. The protocol works in three steps and does not require \textit{any} data of old tasks. First, we split the training set of the current learning step into train and validation sets. We use $80\%$ of the data for training and $20\%$ for validation. Note that the validation set contains only labels for the current learning step. 

Second, we set general \hypers values (\eg learning rate) as the ones achieving the highest accuracy in the new set of classes with the fine-tuned model. Since we tested multiple methods, we wanted to ensure fairness in terms of \hypers used, without producing biased results. To this extent, this step is held out only once starting from the fine-tuned model and fixing the \hypers for all the methods. In particular, we set the learning rate as $10^{-3}$ for the incremental steps in all datasets and settings.

As a final step, we set the \hypers specific of the continual learning method as the highest values (to ensure minimum forgetting) with a tolerated decay on the performance on the new classes with respect to the ones achieved by the fine-tuned model (to ensure maximum learning). We set the tolerated decay as $20\%$ of the original performances, exploring \hypers values of the form $A\cdot 10^{B}$, with $A\in\{1,5\}$ and $B\in\{-3,\dots,3\}$. We perform this validation procedure in the first learning step of each scenario, keeping the \hypers fixed for the subsequent ones. Since this procedure is costly, we perform it only for the Pascal-VOC dataset, keeping the \hypers for the large-scale ADE20k. As a result, for the prior focused methods, we obtain a weight of $500$ for EWC \cite{kirkpatrick2017overcoming} and PI \cite{zenke2017continual} and $100$ for RW \cite{chaudhry2018riemannian} in all scenarios. For the data-focused methods we obtain a weight of $100$ for the distillation loss of LwF \cite{li2017learning}, $10$ for the one in LwF-MC \cite{rebuffi2017icarl} and $100$ for both distillation losses in ILT \cite{michieli2019incremental}, in all settings. For our MiB method, we obtain a distillation loss weight of $10$ for all scenarios except for the \textit{15-1} in Pascal VOC, where the weight is set to $100$. 

% Since this protocol has been defined for continual learning of new tasks (\ie with multiple, separate classifiers), we adapt it to incremental class learning for semantic segmentation by measuring $A_p$ on the set of new classes, excluding the background. Since we tested multiple methods, we wanted to ensure fairness in terms of \hypers used, without producing biased results. To this extent, the choice of the general \hypers (\ie second step) is held out once starting from the fine-tuned model (FT) and fixing the \hypers (\eg learning rate, weight-decay)for all the methods. The used \hypers are detailed in the main paper. For the \hypers specific for each model, we held-out the validation procedure in the first learning step of each scenario, keeping the The \hypers fixed for the subsequent ones. Since the validation procedure is costly, we perform it only for the Pascal-VOC dataset, keeping the found \hypers for the large-scale ADE20k scenario. To summarize, we report in Table \ref{tab:hypers} the \hypers used for all the methods and scenarios.

%* descrizione protocollo validazione
%* descrizione su dove abbiamo fatto validation.
%* Lambdas per ogni metodo.

\section{Code}
Together with the supplementary material, we provide the code used to obtain our results.
The code can be found at \url{https://github.com/fcdl94/MiB}. 
We implement all the methods using the PyTorch\footnote{\url{https://pytorch.org/}} framework, performing our experiments on two NVIDIA Titan RTX GPUs. We provided all the instructions to set up all the incremental learning scenarios of both ADE20k and Pascal VOC, as well as all the requirements for training and testing our method and all the baselines. The instructions are available in the \textit{README.md} file.

{\small
\bibliographystyle{ieee_fullname}
\bibliography{egbib}
}